\pdfoutput=1

\documentclass[11pt]{article}

\usepackage[final]{acl}

\usepackage{times}
\usepackage{latexsym}

\usepackage[T1]{fontenc}

\usepackage[utf8]{inputenc}

\usepackage{microtype}

\usepackage{inconsolata}

\usepackage{graphicx}

\usepackage{times}
\usepackage{latexsym}
\usepackage{algorithm}
\usepackage{amsmath}
\usepackage{xcolor}
\usepackage{colortbl}
\usepackage[T1]{fontenc}

\usepackage[utf8]{inputenc}

\usepackage{microtype}

\usepackage{inconsolata}
\usepackage{multirow}
\usepackage{graphicx}
\usepackage{subfigure}
\usepackage{amsmath}
\usepackage{booktabs}
\usepackage{algorithm}
\usepackage{algpseudocode}
\usepackage{subcaption}
\usepackage[normalem]{ulem}
\usepackage{xcolor}
\definecolor{myblue}{rgb}{0.267, 0.447, 0.769} 
\definecolor{mygreen}{rgb}{0, 0.690, 0.314}
\def\method{\textsc{ThinkNote}}
\title{\textsc{ThinkNote}: Enhancing Knowledge Integration and Utilization of Large Language Models via Constructivist Cognition Modeling}

\author{Zhipeng Xu$^{1}$, Zhenghao Liu$^{1}$\thanks{ \ \ indicates corresponding author.}, Yukun Yan$^{2}$, Shuo Wang$^{2}$, Shi Yu$^{2}$, \\ \textbf{Zheni Zeng$^{2}$, Chaojun Xiao$^{2}$, Zhiyuan Liu$^{2}$, Ge Yu$^{1}$ and Chenyan Xiong$^{3}$} \\ 
$^1$School of Computer Science and Engineering, Northeastern University, Shenyang, China \\
$^2$Department of Computer Science and Technology, Tsinghua University, Beijing, China \\
$^3$Language Technologies Institute, Carnegie Mellon University, Pittsburgh, United States\\
}

\begin{document}

\maketitle
\begin{abstract}
Large Language Models (LLMs) have demonstrated strong performance across a wide range of NLP tasks. However, they often exhibit suboptimal behaviors and inconsistencies when exposed to unfamiliar external information, underscoring their limitations in effectively leveraging such knowledge. 
Inspired by constructivist learning theory, we propose \method{}, a novel framework that enhances the external knowledge utilization of LLMs through a two-stage constructivist cognitive modeling process. Specifically, \method{} performs knowledge assimilation to align new information with the model's parametric memory, forming a coherent internal representation. It then applies thought accommodation to adapt internal reasoning, thereby promoting more consistent and reliable outputs.
Extensive experimental results demonstrate that \method{} achieves a 10\% improvement over strong baseline methods on various question-answering benchmarks. Further analysis indicates that \method{} effectively integrates and utilizes external knowledge to help LLMs generate accurate responses and improve their self-consistency.
All data and codes are available at \url{https://github.com/OpenMatch/ThinkNote}.
\end{abstract}

\section{Introduction}
Large Language Models (LLMs), \textit{e.g.} ChatGPT-4~\cite{openai2023gpt} and LLaMA~\cite{touvron2023llama}, have shown strong emergent abilities and achieved convincing performance in various NLP tasks~\cite{wei2022emergent,zhao2023survey}. 
Despite these advancements, LLMs often face challenges in effectively leveraging external knowledge to generate accurate responses, especially when encountering information that is noisy or incomplete~\cite{liu2023lost,xieadaptive,asai2023self}. 
Such limitations lead to response inconsistencies and hallucinations, constraining their application in knowledge-intensive scenarios~\cite{ji2023survey,xu2024face4rag,wang2023can}.

Recent research investigates strategies for enhancing the integration and utilization of external knowledge in LLMs. 
These efforts aim to develop specialized modules that refine, reflect on, or summarize external knowledge before integration~\cite{yu2023chain, xu2023recomp} or filter the external knowledge based on its relevance to the current context~\cite{peng2023check, zhao2023thrust, asai2023self}.
While these methods show their effectiveness in helping LLMs acquire new knowledge~\cite{ji2023survey,wei2024instructrag}, they still treat LLMs as passive recipients of information~\cite{steffe1995constructivism,larochelle1998constructivism} and fail to address the underlying cognitive limitations that lead to insufficient knowledge utilization~\cite{yu2025unveiling}.
This limitation leads to incorrect responses and can even cause degraded responses on certain tasks~\cite{foulds2024ragged,shuster2021retrieval}.

To address these limitations, we propose a shift from treating LLMs as passive consumers of knowledge to modeling them as active knowledge constructors, drawing inspiration from constructivist cognitive theories~\cite{steffe1995constructivism, papert1980children, piaget1972psychology}.
Constructivism emphasizes that knowledge is not simply absorbed from the environment, but is actively built through experience, reflection, and assimilation of new information into existing mental frameworks~\cite{von1984constructivism}. 
Under this paradigm, external knowledge is not directly fused into the input context of LLMs but transformed and recontextualized based on the model's current understanding of the task or question~\cite{schunk2012learning}.
Such a cognitively grounded design encourages models to resolve inconsistencies, discard irrelevant or misleading information, and synthesize new insights, similar to how humans build reliable understanding from noisy data~\cite{bruner1990acts}.

In this paper, we propose \method{}, a cognition-inspired framework to activate LLMs to effectively integrate and utilize external knowledge by mimicking the cognitive processes of assimilation and accommodation from constructivism~\cite{steffe1995constructivism}. 
Specifically, \method{} incorporates knowledge assimilation to transform external information into coherent insights and understanding, extending beyond query-specific notes~\cite{yu2023chain} to enrich the parametric knowledge of LLMs. The thought accommodation then updates the parametric memory of LLMs by incorporating the assimilated knowledge to refine and enhance the internal reasoning chain of LLMs. 
This two-step cognitive process is prompted to mimic different learning behaviors of humans, where we configure \method{} using four instructions to process external knowledge by associating familiar knowledge, anchoring unfamiliar information, extracting logical reasoning, or identifying counterfactuals. 

Our experiments on a range of knowledge-intensive question-answering tasks demonstrate the effectiveness of \method{}.
Compared with directly incorporating external knowledge, \method{} consistently achieves around a 10\% performance improvement across backbone LLMs of different scales.
Moreover, \method{} exhibits stronger robustness when facing noisy, incomplete, or partially misleading external knowledge, suggesting that actively constructing and reorganizing knowledge is more reliable than passive conditioning on retrieved evidence. 
Further analysis reveals that \method{} leads to more concentrated information accumulation~\cite{fan2025improving} on a small set of salient tokens, indicating its ability to identify and leverage critical elements of external knowledge during response generation.

\section{Related Work}
Large Language Models (LLMs) consistently face stability challenges when utilizing external knowledge~\cite{zhao2023thrust,asai2023self}. 
While LLMs are exposed to extensive corpora during training and demonstrate strong generalization across various tasks~\cite{zhao2023survey,touvron2023llama}, their performance declines when encountering unfamiliar, ambiguous, or contradictory information~\cite{ji2023survey, liu2023lost, foulds2024ragged}.  
These limitations illustrate a notable gap between the availability of external knowledge and the model's capacity to effectively integrate and apply it, especially evident in knowledge-intensive tasks~\cite{yu2023augmentation,shi2023replug,jiang2023active}.

To mitigate these limitations, some methods attempt to enhance knowledge integration by optimizing the interface between external information and the model's input context~\cite{cuconasu2024power,gao2023retrieval}. 
These methods typically aim to filter or reformat noisy inputs through knowledge summarization~\cite{xu2023recomp}, reflection~\cite{asai2023self}, or multi-step verification~\cite{yu2023chain, peng2023check}.
While these approaches have advanced the ability of LLMs to integrate and utilize external information, they often push the burden of knowledge alignment to external modules~\cite {wang2024astute}, where the LLMs remain conceptualized as passive processors, failing to reorganize or evaluate new information based on their current understanding~\cite{madaan2023self}. This design choice limits their ability to support adaptive knowledge integration, especially in scenarios that require the model to actively construct and revise its understanding.

More recent efforts explore internal modifications to how LLMs represent, revise, or adapt knowledge. 
These include work on memory editing~\cite{meng2022locating,trivedi2023interleaving}, belief tracking~\cite{wilie2024belief}, and metacognitive strategies such as self-verification and response calibration~\cite{madaan2023self,liu2023lost}. 
While promising, these approaches are often fragmented and tool-like: they provide mechanisms to intervene in model behavior without modeling the interaction between the model and external knowledge~\cite{collins2024building}.
There is still limited consensus on what constitutes effective knowledge integration at the cognitive level, and most current methods operate under engineering-driven constraints rather than principles drawn from learning theory or epistemology~\cite{floridi2024anthropomorphising}.

\section{Methodology}
In this section, we introduce our \method{} framework, grounded in constructivist cognitive theory~\cite{steffe1995constructivism}, which aims to emulate the cognitive process of knowledge learning employed by human learners to interact with external knowledge. 
We first describe the overall workflow of \method{} framework (Sec.~\ref{model:workflow}) and then elaborate on the specific configuration strategies used to regulate learning behavior (Sec.~\ref{model:role}).

\begin{figure*}
    \centering
    \includegraphics[width=\linewidth]{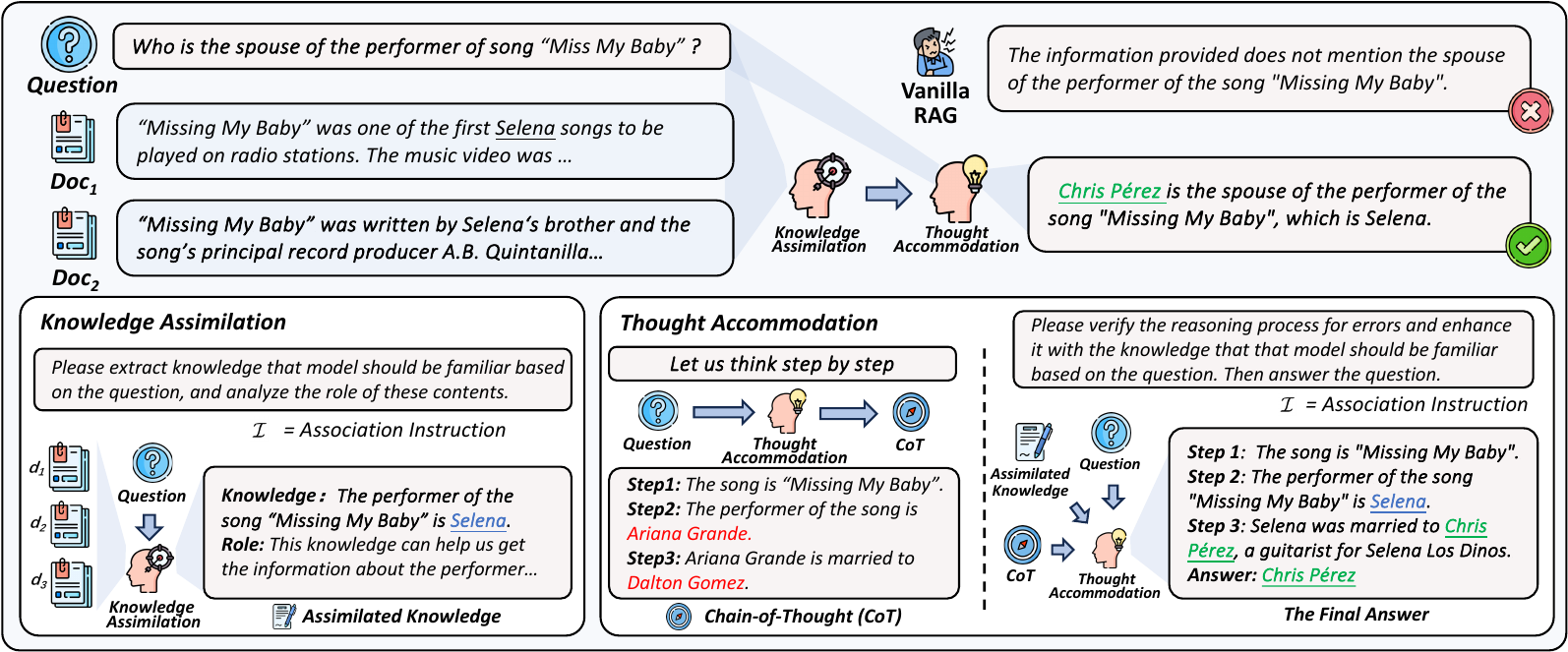}
    \caption{The Overview of the \method{} Workflow. The framework follows a two-stage process of knowledge assimilation and thought accommodation to integrate external knowledge into model reasoning.}
    \label{fig:model}
\end{figure*}

\subsection{Enhancing Knowledge Integration and Utilization in LLMs via \method{}}
\label{model:workflow}
As shown in Figure~\ref{fig:model}, \method{} is a cognition-inspired framework designed to enhance the ability of LLMs to integrate and utilize external knowledge to generate accurate responses, which draws on the cognitive principles of constructivist learning theory~\cite{steffe1995constructivism}.
This framework simulates a two-stage cognitive process:
\method{} first conducts knowledge assimilation to align new information with the model's parametric memory, and then applies
thought accommodation to adapt internal reasoning, thereby promoting more consistent and reliable outputs.

\textbf{Knowledge Assimilation.} 
Humans activate relevant memory (which is referred to as schema in constructivist cognitive theory) to interpret the meaning and function of external information and construct new understanding, which is called knowledge assimilation~\cite{steffe1995constructivism, moreno1999cognitive}.
The knowledge assimilation stage in \method{} simulates this process, driving the LLMs to align and integrate external knowledge with their internal parametric memory and point out the function of knowledge for problem-solving.
Specifically, given a query $q$ and a set of external knowledge $\mathcal{D}$, \method{} guides the foundational model $\mathcal{M}$ to generates an assimilated rationale $\mathcal{T}$, which can be expressed as:
\begin{equation}\label{eq:ka_agent}
    \mathcal{T} = \mathcal{M}\left(\mathcal{I}_\text{ka}, q, \mathcal{D}\right),
\end{equation}
where $\mathcal{I}_{ka}$ indicates the instruction to mimic different learning behaviors. The assimilated knowledge $\mathcal{T}$ extends beyond knowledge summarization~\cite{yu2023chain} and serves to contextualize the knowledge within the problem scenarios, enabling the model to not only recall relevant facts but also explain how and why these facts matter. 

\textbf{Thought Accommodation.} 
The thought accommodation refers to the adjustment of internal memory when new information challenges prior understanding in constructivist theory~\cite{steffe1995constructivism,larochelle1998constructivism}. 
In \method{}, this process is simulated by guiding the model $\mathcal{M}$ to revise its initial reasoning based on assimilated knowledge $\mathcal{T}$.
Given a query $q$, \method{} first conducts self-inquiry to generate a preliminary chain-of-thought $\mathcal{R}$ using its parametric memory:
\begin{equation}
\mathcal{R} = \mathcal{M}\left(\mathcal{I}_\text{CoT}, q\right).
\end{equation}
Here $\mathcal{I}_\text{CoT}$ triggers a chain-of-thought reasoning grounded in prior knowledge, which may lack context or accuracy due to outdated or incomplete parametric memory.
To address this, \method{} guides the model $\mathcal{M}$ to revise the chain-of-thought $\mathcal{R}$ using the assimilated knowledge $\mathcal{T}$, producing the final output as:
\begin{equation}\label{eq:ta_agent}
y = \mathcal{M}\left(\mathcal{I}_{ta}, q, \mathcal{T}, \mathcal{R}\right),
\end{equation}
where $\mathcal{I}_{ta}$ instructs the model $\mathcal{M}$ to align its reasoning with assimilated external knowledge. 
Through this thought accommodation process, \method{} enables the model to approximate cognitive disequilibrium regulation, detecting mismatches between external knowledge and internal reasoning.

\subsection{Configuring \method{} by Mimicking Different Human Learning Behaviors}\label{model:role}
Learning behaviors are goal-directed mental operations used to enhance knowledge utilization and problem-solving.
\method{} employs flexible instructions $\mathcal{I}_{ka}$ and $\mathcal{I}_{ta}$ to simulate diverse human learning behaviors at each stage of the cognitive process, supporting both knowledge assimilation and thought accommodation.
The instruction is instantiated in one of four constructivist modes: Anchoring, Association, Reasoning, and Reflection.
The details of these instruction templates are shown in Appendix~\ref{app:prompts}.

\textbf{Association.} 
The association behavior maps new facts to known entities. 
Rather than introducing novel concepts, it reinforces and elaborates on familiar knowledge. 
For instance, when encountering information about electric vehicles, the model relates it to existing knowledge in their parametric memory of ``batteries'' and ``charging stations'', thereby deepening its understanding by associating these new contents to previously learned concepts~\cite{steffe1995constructivism}.

\textbf{Anchoring.} 
When humans encounter a new concept like ``photosynthesis'' in the query, they would start by identifying more basic concepts such as ``chlorophyll'' or ``carbon dioxide'' in external knowledge to create reference points in memory. 
Similarly, the instruction of anchoring behavior guides models to simulate this behavior and help them form an initial understanding by extracting related information from unfamiliar external knowledge.
This behavior grounds its comprehension of queries and facilitates the expansion of topic-relevant knowledge~\cite{yu2023chain}.

\textbf{Reasoning.} Inspired by~\citet{edge2024local}, the reasoning behavior guides the model to extract and organize logical relationships found in the external knowledge. 
This enables LLMs to build coherent and structured explanations of external knowledge. 
For example, when synthesizing documents on climate change policy, the model infers causal links, such as ``carbon tax leads to emission reduction, which in turn impacts long-term economic growth''---to support principled reasoning.

\textbf{Reflection.} 
This behavior encourages the model to evaluate its outputs by comparing internal knowledge with external knowledge. 
By treating the external documents as authoritative references, the model can detect inconsistencies and revise its responses. For instance, if the model initially generates an outdated population statistic, reflection allows it to identify and correct the discrepancy based on up-to-date knowledge~\cite{xu2024hallucination}.

\section{Experimental Methodology}
In this section, we describe the datasets, baselines, and implementation details of our experiments.

\textbf{Dataset.} Our experiment uses six datasets to evaluate the performance of \method{}, including NQ~\cite{kwiatkowski2019natural}, PopQA~\cite{mallen2023not}, TriviaQA~\cite{joshi2017triviaqa}, 2WikiMultiHopQA~\cite{ho2020constructing} and ASQA~\cite{stelmakh-etal-2022-asqa}. 
We use Contriever~\cite{izacard2021unsupervised} to obtain the external knowledge for all datasets to ensure fairness, following previous work~\cite{asai2023self,wei2024instructrag}. Due to the cost of inference, we randomly sample a subset of 500 questions from each dataset in our experiments in line with prior work~\cite{trivedi2023interleaving,yoran-etal-2023-answering}.

\begin{table*}[!t]
\centering
\small
\begin{tabular}{l|l|cccccccc}
\toprule
\multirow{2}{*}{\textbf{Method}} & \multirow{2}{*}{\textbf{LLM}} & 
PopQA & NQ & TriviaQA  & \multicolumn{2}{c}{2WikiMHQA} & \multicolumn{3}{c}{ASQA} \\ 
& & (\textit{acc}) & (\textit{acc}) & (\textit{acc}) & (\textit{acc}) & (\textit{rec}) & (\textit{str-em}) & (\textit{hit}) & (\textit{rec})  \\ 
\midrule
\multicolumn{10}{l}{{\cellcolor[rgb]{0.957,0.957,0.957}}\textbf{Vanilla LLMs}} \\
\midrule
\multirow{3}{*}{Direct QA} & Llama-3-Ins$_{\textsc{8b}}$  
& 24.2 & 39.2 & 67.2 & 43.0 & 48.1 & 24.8 & 5.0 & 16.1\\
& Llama-3-Ins$_{\textsc{70b}}$  
& 34.2 & 54.2 & 80.4  & 53.8 & 60.0 & 33.3 & 8.8 & 20.2\\
& ChatGPT-4o$_{\textsc{Mini}}$
& 32.6 & 51.0 & 75.0  & 47.4 & 52.3 & 31.4 & 7.8 & 18.5\\ \midrule
\multirow{3}{*}{CoT} & Llama-3-Ins$_{\textsc{8b}}$  
& 24.8 & 44.0 & 69.4  & 47.2 & 51.4 & 28.8 & 7.8 & 25.5\\
& Llama-3-Ins$_{\textsc{70b}}$  
& 31.6 & 54.4 & 80.6  & 56.4 & 62.6 & 36.4 & 11.2 & 32.7\\
& ChatGPT-4o$_{\textsc{Mini}}$ & 32.4 & 53.2
& 76.2 & 51.0 & 55.3 & 32.4 & 8.0 & 21.6\\
\midrule
\multicolumn{10}{l}{{\cellcolor[rgb]{0.957,0.957,0.957}}\textbf{LLM w/ RAG}} \\\midrule
\multirow{3}{*}{Vanilla RAG} & Llama-3-Ins$_{\textsc{8b}}$  
& 59.8 & 56.0 & 71.2 & 45.2 & 49.3 & 34.2 & 12.0 & 27.7\\
& Llama-3-Ins$_{\textsc{70b}}$  
& 63.4 & 59.8 & 75.2 & 52.2 & 56.3 & 38.1 & 14.6 & 29.8\\
& ChatGPT-4o$_{\textsc{Mini}}$ 
& 62.0 & 64.8 & 77.8 & 51.4 & 56.2 & 38.9 & 13.2 & 28.0\\ \midrule
\multirow{4}{*}{Self-Refined}
& Self-RAG$_{\textsc{7b}}$
& 53.8 & 43.4 & 66.4 & 36.6 & 40.2 & 28.6 & 9.0 & 13.2 \\
& Self-RAG$_{\textsc{8b}}$
& 54.8 & 46.2 & 67.0 & 37.2 & 42.6 & 37.6 & 15.6 & 21.4\\
& Chain-of-Note$_{\textsc{8b}}$
& 64.2 & 59.8 & 73.0 & 52.0 & 56.2 & 42.0 & 17.8 & 56.2\\
& Chain-of-Note$_{\textsc{70b}}$
& 68.2 & 67.8 & 78.4 & 56.8 & 61.1 & 45.0 & 20.6 & 54.9 \\ \midrule
\multirow{3}{*}{\method{}} & Llama-3-Ins$_{\textsc{8b}}$  
& 65.8 & 62.0 & 79.8 & 53.8 & 60.0 & 43.4 & 18.8 & 57.8\\
& Llama-3-Ins$_{\textsc{70b}}$  
& \textbf{69.8} & 68.4 & \textbf{85.4} & \textbf{63.2} & \textbf{68.4} & 48.8 & 22.6 & 59.9\\
& ChatGPT-4o$_{\textsc{Mini}}$
& 69.8 & \textbf{71.0} & 83.4 & 61.0 & 67.4 & \textbf{51.5} & \textbf{24.6} & \textbf{61.3}\\

\bottomrule
\end{tabular}
\caption{Overall Performance of \method{} and Baselines on Different Knowledge-Intensive Benchmarks. For \method{}, we use the instruction of association learning behavior to guide both knowledge assimilation and thought accommodation and evaluate the overall performance. 
    \label{tab:main}} 
\end{table*}
\textbf{Evaluation Metrics.} 
Following \citet{asai2023self} and \citet{wei2024instructrag}, we use \textit{correctness} (\textit{str-em}), \textit{hit rates} (\textit{hit}), and \textit{recall} (\textit{rec})  to evaluate the question-answering performance on the ASQA task and use \textit{accuracy} (\textit{acc}) for other tasks. All of the evaluation metrics are implemented referring to the ALCE toolkit~\cite{gao2023enabling}.

\textbf{Baselines.} 
The baselines consist of vanilla LLMs, LLMs with CoT, and some RAG models. 
We use Meta-Llama-3-Instruct and ChatGPT-4o-Mini as backbone LLMs to implement baseline models. 
Detailed descriptions, configurations, and implementation specifics of each baseline model are provided in Appendix~\ref{app:baseline}.

\textit{Vanilla LLMs.} We employ two methods to prompt LLMs to answer questions based solely on their parametric memory. First, the Direct QA approach involves directly instructing the LLMs to answer questions. Second, we implement the Chain-of-Thought~\cite{wei2022chain} prompting, which generates a step-by-step reasoning process to arrive at the answer.

\textit{RAG Models.}  We implement three widely used RAG models as baselines, including vanilla RAG, Chain-of-Note~\cite{yu2023chain}, and Self-RAG~\cite{asai2023self}. The vanilla RAG model directly feeds retrieved passages as the context and asks the LLM to generate the answer~\cite{ram2023context}. Chain-of-Note extends the CoT method to summarize the query-related knowledge from external knowledge for producing answers. Self-RAG~\cite{asai2023self} trains LLMs to employ a self-reflection mechanism to filter out irrelevant evidence effectively.

\textbf{Implementation Details.}
We use different language models, including Meta-Llama-3-Instruct-8B, Meta-Llama-3-Instruct-70B, and ChatGPT-4o-Mini as the foundation model to evaluate the performance of the \method{}. Specifically, we deploy Meta-Llama-3-Instruct models using the vLLM~\cite{kwon2023efficient}, and use the OpenAI SDK to call the ChatGPT-4o-Mini API for experiments. The temperature is set to 0.2 for all models. 

\section{Evaluation Result}
In this section, we first present the overall performance of \method{}. We then analyze its effectiveness in leveraging knowledge. Finally, we demonstrate how \method{} enhances the self-consistency of LLMs in answering questions.

\subsection{Overall Performance}
The overall performance of \method{} on knowledge-intensive tasks is shown in Table~\ref{tab:main}. 

Compared to vanilla RAG models, \method{} delivers over a 10\% improvement in performance, demonstrating its effectiveness in enabling LLMs to more accurately integrate and utilize external knowledge for question answering. 
While self-refinement methods focus on denoising external information to enhance generation, \method{} surpasses these approaches with over a 4\% improvement,  highlighting the importance of fostering a deeper understanding of knowledge within LLMs rather than merely extracting surface-level content related to queries.
Furthermore, \method{} consistently outperforms all baseline models, regardless of the model scale or whether the backbone is black or white-box, indicating strong generalization and robustness across different model configurations.

Compared with vanilla RAG models, the Self-Refined methods exhibit performance degradation on TriviaQA and 2WikiMHQA datasets, showing that filtering query-related contents may risk omitting critical information from external knowledge.
Unlike Self-Refined RAG, \method{} employs a two-stage cognitive process that actively acquires essential knowledge from external information, rather than relying solely on LLMs to evaluate the relevance between queries and evidence. 
By leveraging constructivist cognitive modeling, \method{} offers a promising way to emulate human learning behaviors and demonstrates its effectiveness by consistently achieving improvements across all test scenarios.

\begin{table*}[t]
\centering
\small
\begin{tabular}{l|ccc|ccc}
\toprule
\multirow{3}{*}{\textbf{Method}} & \multicolumn{3}{c|}{\textbf{Llama-3-Ins-8B}}& \multicolumn{3}{c}{\textbf{Llama-3-Ins-70B}}\\
& PopQA & 2WikiMHQA & ASQA  & PopQA & 2WikiMHQA & ASQA\\
 & (\textit{acc}) & (\textit{acc}) & (\textit{str-em})  & (\textit{acc}) & (\textit{acc}) & (\textit{str-em}) \\
\midrule
\rowcolor{gray!8}\multicolumn{7}{l}{\textbf{Cognitive Modeling Ablation}} \\\midrule
\method{} & {65.8}  & {53.8} & {43.4} & {69.8} & {63.2} & {48.8}  \\
{w/o \textit{Knowledge Assimilation}} & 58.4 & 48.2 & 38.6 & 61.8 & 59.6 & 43.2 \\
{w/o \textit{Thought Accommodation}} & 63.0 & 46.8 & 40.2 & 66.0 & 57.4 & 45.7 \\
\midrule
\rowcolor{gray!10}\multicolumn{7}{l}{\textbf{Learning Behaviors Ablation}} \\\midrule
{w. Anchoring Instruction} & 64.8 & \textbf{54.6} & 39.9 & 69.0 & \textbf{63.6} & 45.8 \\
{w. Association Instruction} & \textbf{65.8}  & 53.8 & \textbf{43.4} & \textbf{69.8} & 63.2 & 48.8  \\
{w. Reasoning Instruction} & 64.0  & 52.8 & 40.3 & 68.4 & 59.6 & 48.3\\
{w. Reflection Instruction} & 63.2 & 52.8 & 42.2 & 65.2 & 60.6 & \textbf{50.2}\\
\bottomrule
\end{tabular}
\caption{Ablation Study of \method{}.
We examine the contributions of knowledge assimilation and thought accommodation, and vary instruction designs to simulate different learning behaviors. Experiments are conducted with Meta-Llama-3-Instruct-8B and Meta-Llama-3-Instruct-70B.\label{tab:ablation}}
\end{table*}
\subsection{Ablation Study}
\label{main:ablation}
As shown in Table~\ref{tab:ablation}, we conduct ablation studies to evaluate the contribution of each cognitive stage in \method{}.
We further analyze the effectiveness of \method{} under different configurations of knowledge learning behavior instructions.

We first evaluate the performance of \method{} variants on three QA tasks: PopQA, 2WikiMHQA, and ASQA.
For PopQA, the knowledge assimilation stage in \method{} contributes more significantly to performance improvements than the thought accommodation stage, highlighting that answering questions about less frequent entities heavily depends on acquiring information from external knowledge. 
In contrast, for 2WikiMHQA and ASQA, both the knowledge assimilation and thought accommodation modules are equally critical to the effectiveness of \method{}, emphasizing the advantages of chain-of-thought reasoning in tackling more complex QA tasks. 
With increasing model scale, the effectiveness of knowledge assimilation declines, while the advantages of thought accommodation remain steady.
This observation implies that although larger LLMs encode more knowledge within their parameters, deliberate reasoning remains essential for effectively tackling these multi-hop QA and long-form QA tasks.

We next examine the impact of knowledge learning behaviors in \method{} by guiding the knowledge assimilation and thought accommodation stages using different sets of instructions, $\mathcal{I}_{ka}$ and $\mathcal{I}_{ta}$, respectively.
Among all instruction types, the association learning behavior consistently achieves superior performance across diverse QA tasks. This behavior encourages LLMs to reinforce and extend their internal knowledge representations, rather than relying primarily on the assimilation of unfamiliar information from retrieved documents, as is characteristic of anchoring learning behavior.
These findings emphasize the importance of integrating external knowledge through associations with the model's parametric memory. 
For subsequent experiments, we adopt the association learning behavior instruction to configure \method{}.

\begin{figure}[t]
    \centering

    \subfigure[Token-level Information Flow of Vanilla RAG.]{ \label{fig:ifrag} 
    \includegraphics[width=0.48\linewidth]{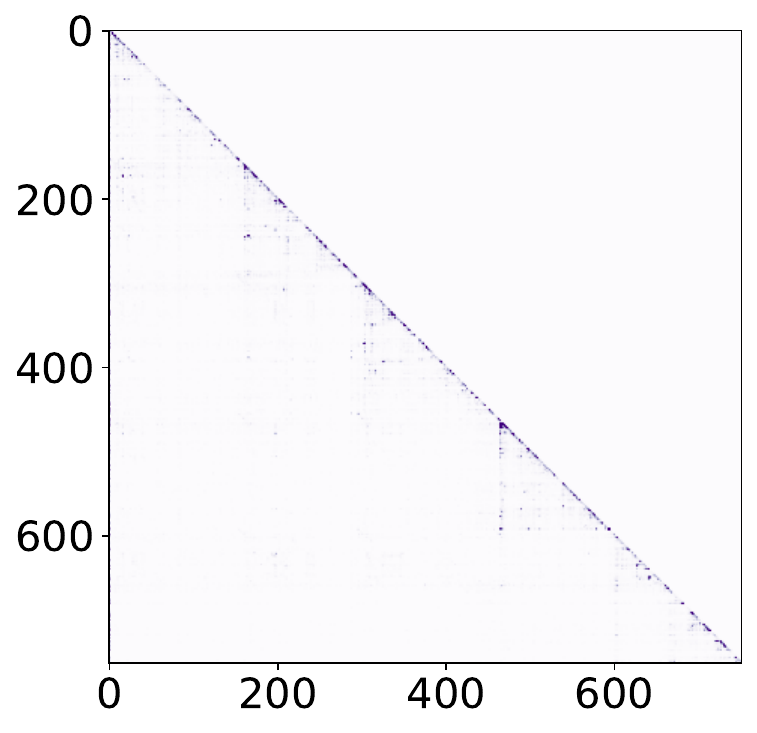}}
    \subfigure[Token-level Information Flow of \method{}.]{ \label{fig:ifarag} 
    \includegraphics[width=0.48\linewidth]{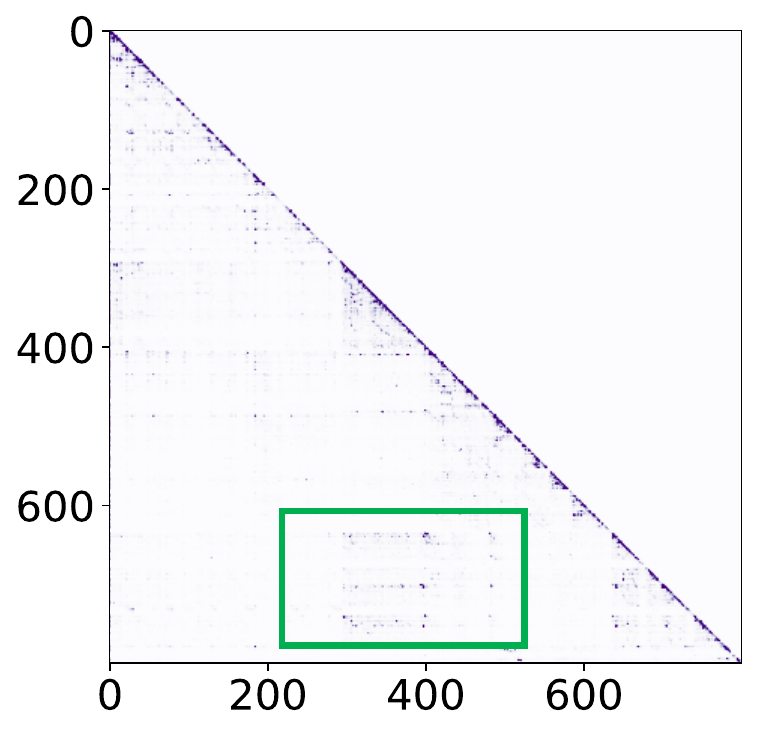}}
    \subfigure[Top Tokens in Information Accumulation as Identified by Vanilla RAG.]{ \label{fig:ss_rag} 
    \includegraphics[width=0.48\linewidth]{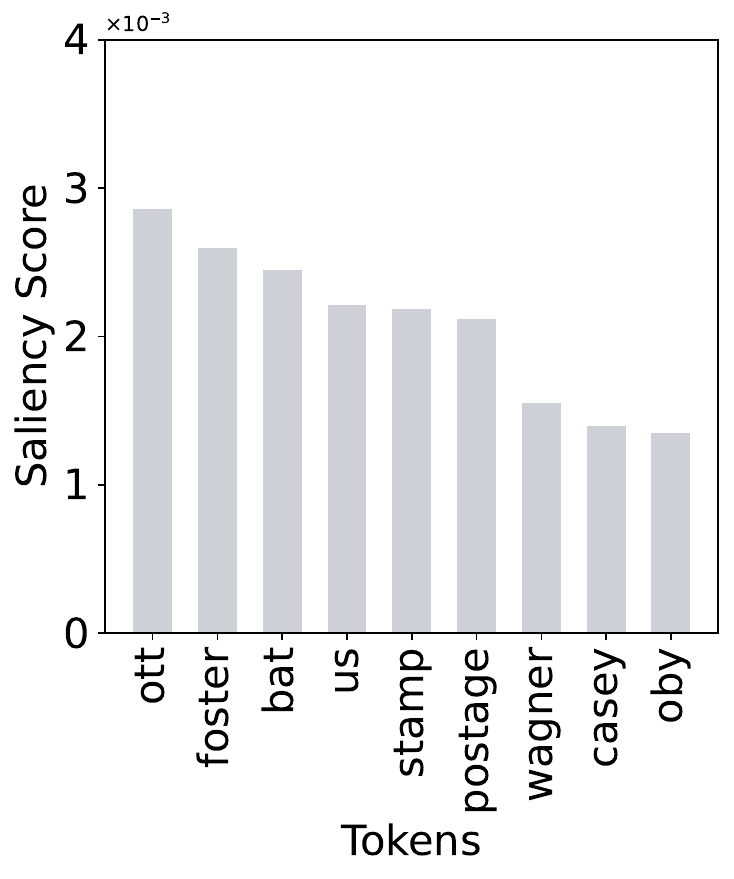}}
    \subfigure[Top Tokens in Information Accumulation as Identified by \method{}.]{ \label{fig:ss_arag} 
    \includegraphics[width=0.48\linewidth]{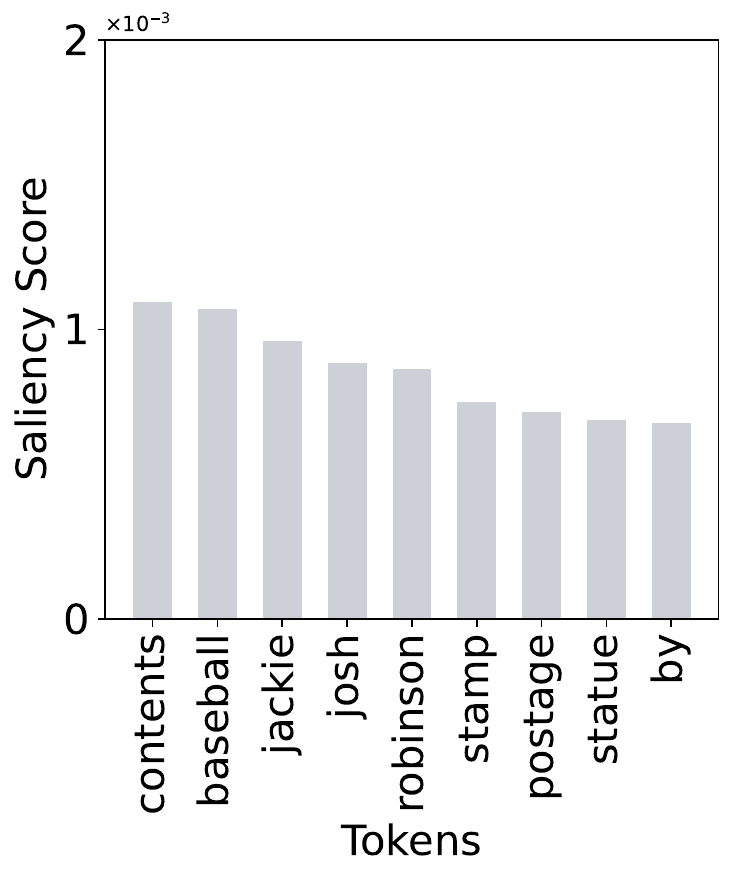}}
    \caption{Information Flow Analysis for \method{}. We implement both Vanilla RAG and \method{} with Meta-Llama-3-Instruct-8B. Darker-colored points indicate higher saliency scores in information flow analysis.}
    \label{fig:ssss}
\end{figure}
\subsection{Information Flow Analysis for \method{} in Knowledge Utilization}
We further analyze how knowledge assimilation and thought accommodation influence the reasoning process in Figure~\ref{fig:ssss}.
Specifically, we sample a case from the NQ dataset and follow \citet{fan2025improving} in utilizing the saliency score to examine interactions between tokens, which is also referred to as information flow. Further details on the saliency score are provided in Appendix~\ref{app:ss}.

We show the information flow for both Vanilla RAG and \method{} in Figure~\ref{fig:ifrag} and Figure~\ref{fig:ifarag}, respectively.
The Vanilla RAG method demonstrates a dense and intricate pattern of information flow, where a broad range of tokens indiscriminately influence other tokens throughout the sequence.
This indicates that Vanilla RAG lacks the mechanism to integrate knowledge or suppress irrelevant information, leading to an over-reliance on incidental associations.
In contrast, \method{} guides the information flow to some specific tokens (highlighted with the green box) by modeling cognitive processes.
These tokens contribute to the accumulation of information flow as part of a chain-of-thought.
This indicates that \method{} can effectively integrate and utilize information to help LLMs generate responses, reflecting a more goal-directed reasoning process, where irrelevant or distracting information is attenuated. 

We further analyze the accumulation of information flow in both Vanilla RAG and \method{} in Figure~\ref{fig:ss_rag} and Figure~\ref{fig:ss_arag}, highlighting the top 10 tokens with the highest accumulated saliency scores. In this example, the query is:
``Who was the first baseball player to be featured on a postage stamp?''
The correct answer is ``Jackie Robinson''.
For Vanilla RAG, the tokens with the highest accumulated saliency are generic terms like ``stamp'' or ``postage''.
While these words are related to the question, they are helpless in answering the question. 
In contrast, \method{} assigns higher saliency to key content words such as ``jackie'', ``robinson'', and ``baseball'', indicating a more effective focus on the discriminative elements for problem-solving. This highlights its effectiveness in gathering useful information to enhance the knowledge utilization capability of LLMs.

\begin{figure}[t]
    \centering
    \subfigure[Retrieval Score (PopQA).]{ \label{fig:popqars} 
    \includegraphics[width=0.48\linewidth]{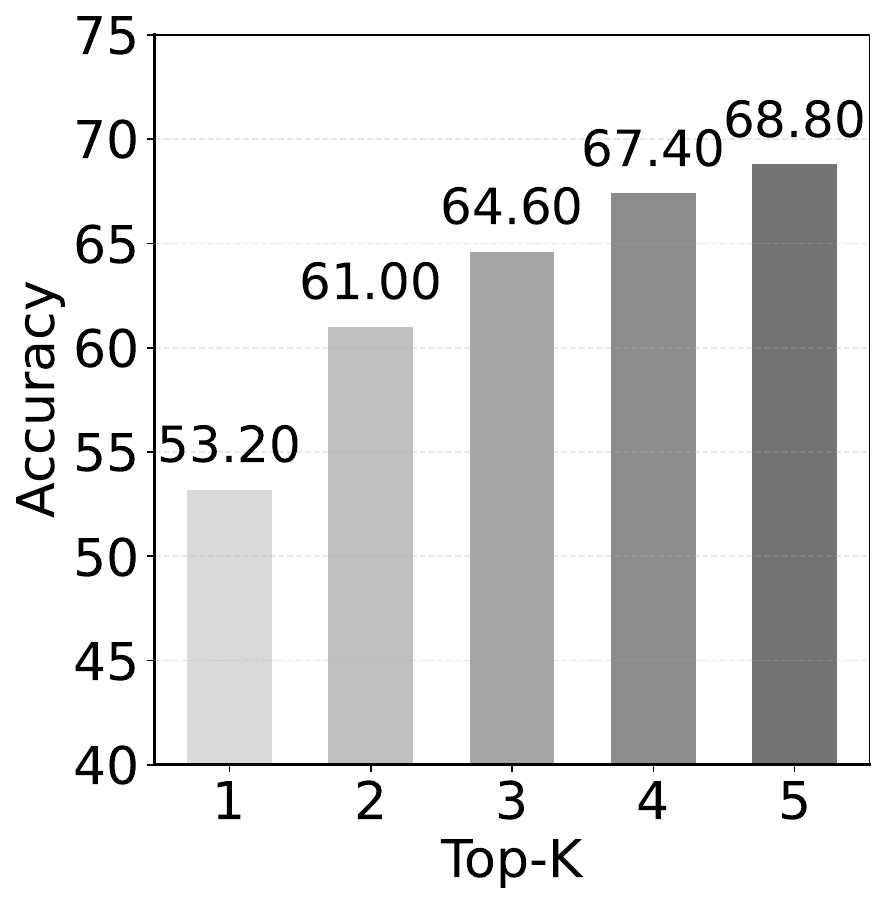}}
    \subfigure[Retrieval Score (NQ).]{ \label{fig:nqrs} 
    \includegraphics[width=0.48\linewidth]{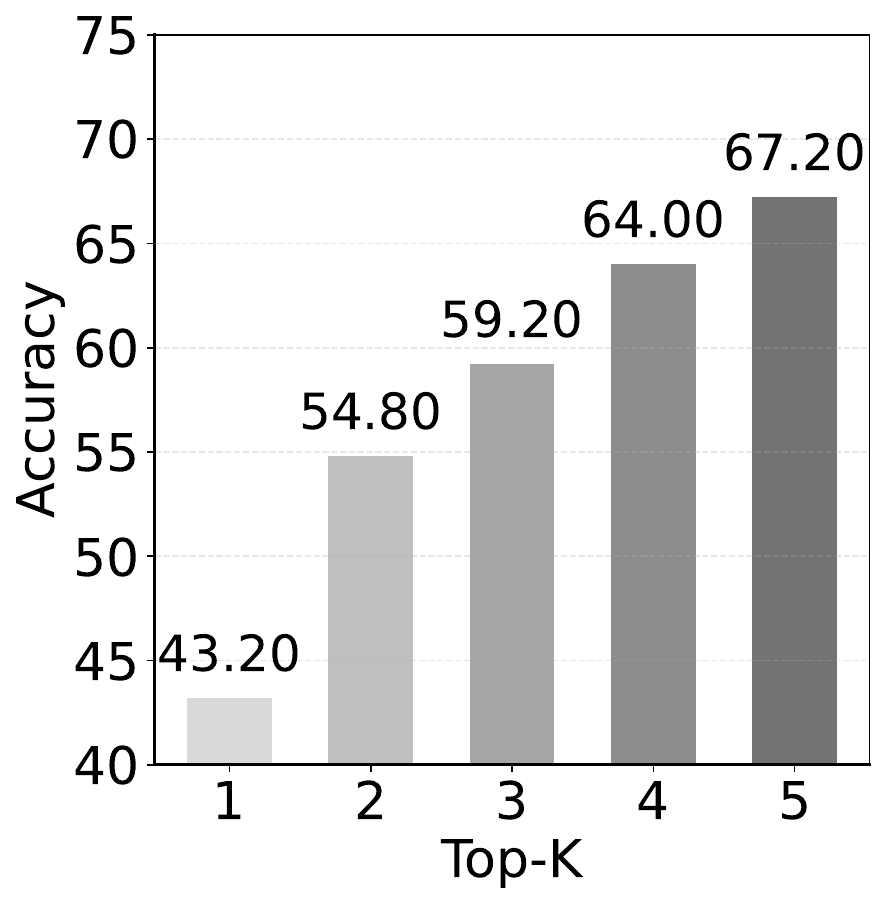}}
    \subfigure[Accuracy (PopQA).]{ \label{fig:popqa} 
    \includegraphics[width=0.48\linewidth]{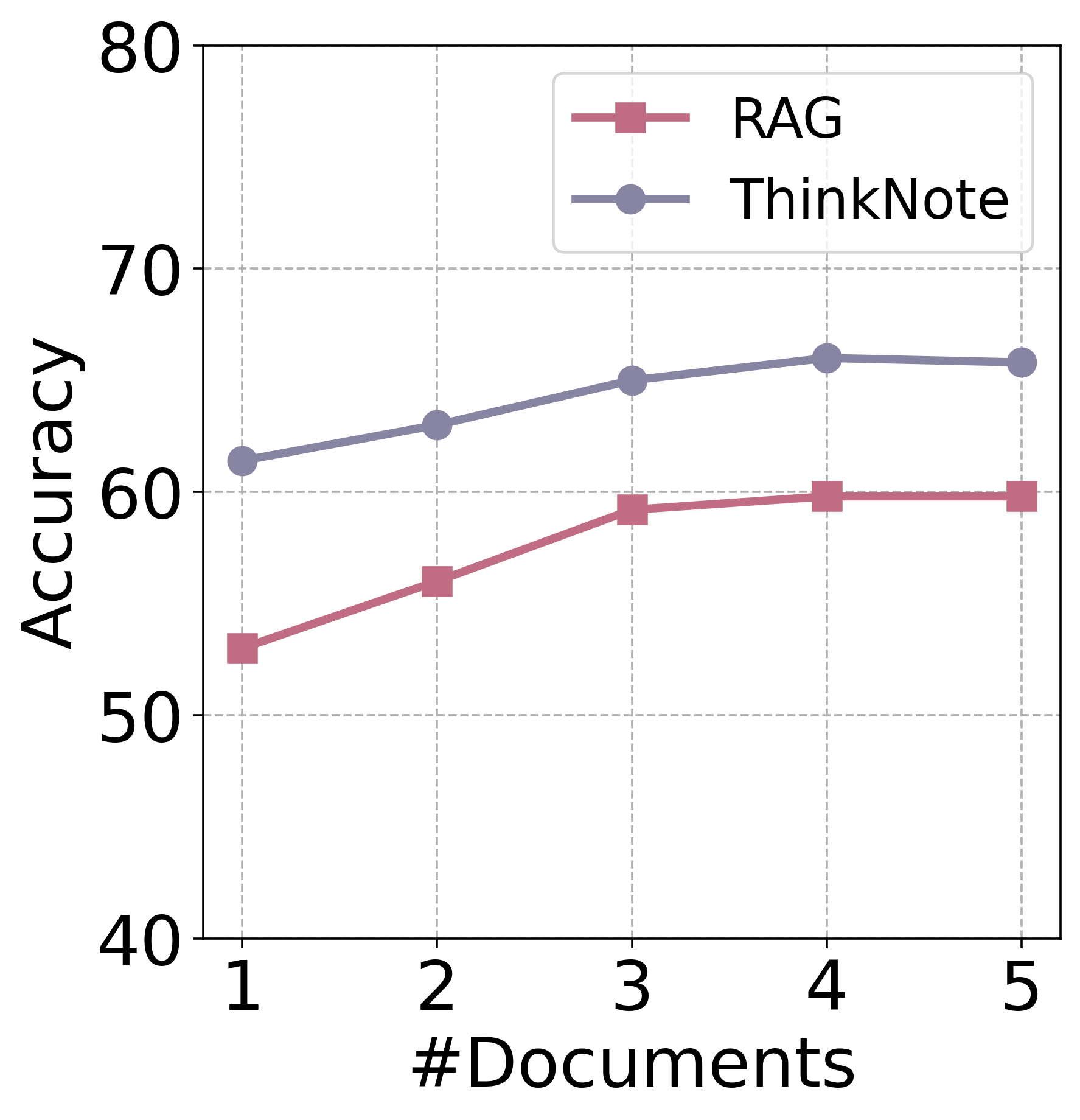}}
    \subfigure[Accuracy (NQ).]{ \label{fig:nq} 
    \includegraphics[width=0.48\linewidth]{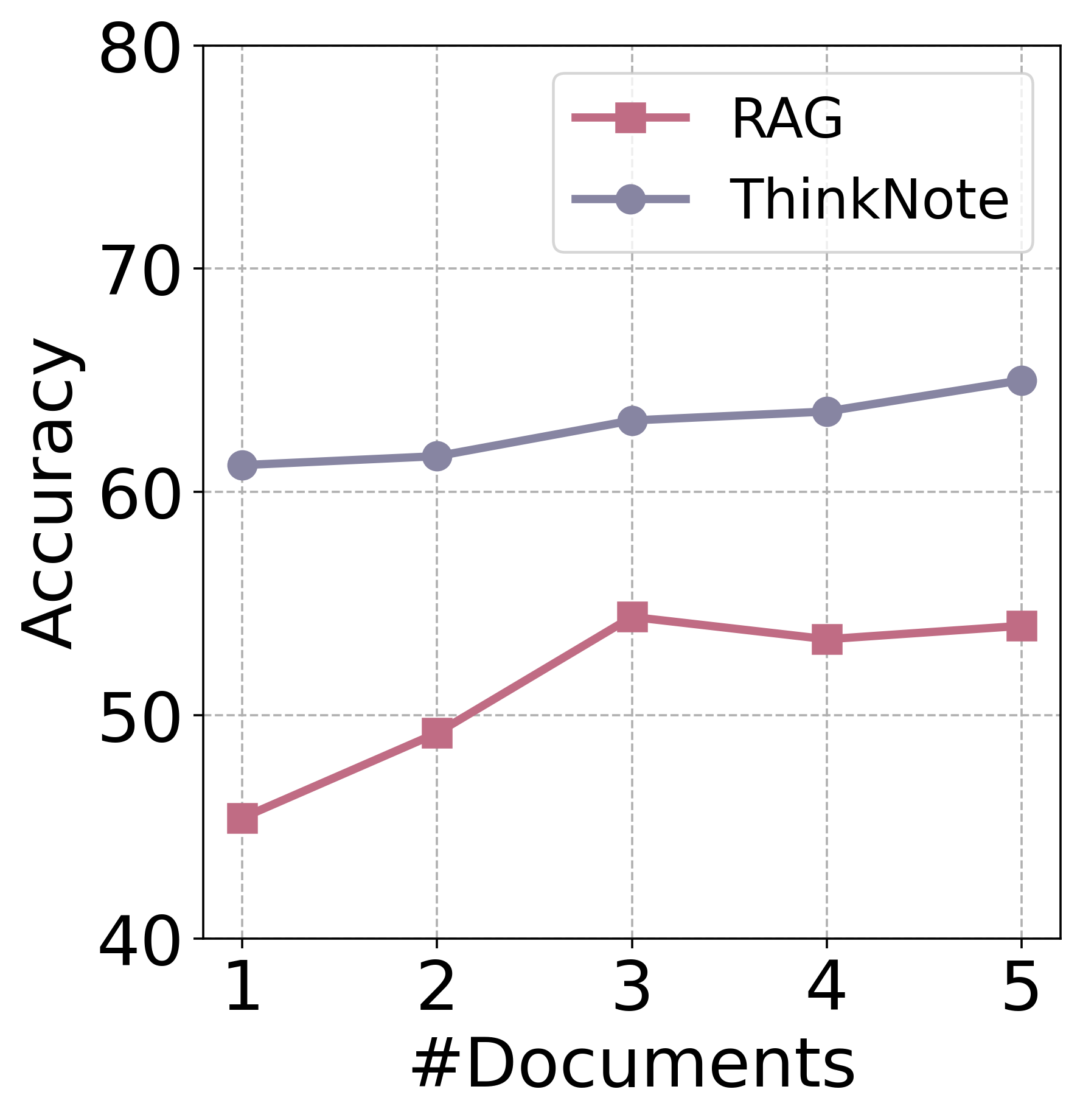}}
    \caption{Performance of RAG and Retrieval Models with Top-$K$ Document Truncation. We implement both Vanilla RAG and \method{} with Meta-Llama-3-Instruct-8B.}
    \label{fig:noise}
\end{figure}
\begin{table*}[t]
\centering
\small
\begin{tabular}{l|ccc|ccc|ccc}
\toprule
\multirow{2}{*}{\textbf{Method}} & \multicolumn{3}{c|}{\textbf{Has-Answer}} & \multicolumn{3}{c|}{\textbf{Miss-Answer}} & \multicolumn{3}{c}{\textbf{Internal Knowledge}}\\ 
& PopQA & NQ  & TriviaQA & PopQA & NQ  & TriviaQA & PopQA & NQ & TriviaQA\\
\midrule
\rowcolor{gray!9}\multicolumn{10}{l}{\textbf{Llama-3-Ins-8B}}  \\
\midrule
Direct QA & 34.0 & 47.5 & 80.1  & 2.6 & 21.4 & 33.8 & 100.0 & 100.0 & 100.0 \\
Vanilla RAG & 85.5 & 76.2 & 91.7 & 3.2 & 6.3 & 18.0 & 89.3 & 79.1 & 88.7  \\ 
Chain-of-Note & 93.9 & 85.6 & 95.0 & 3.8 & 8.2 & 15.8 & 95.9 & 83.2 & 89.0 \\ 
\method{} & 90.1 & 80.4 & 97.0 & 8.3 & 25.8 & 36.0 & 99.2 & 92.9 & 97.3 \\ \midrule
\rowcolor{gray!9}\multicolumn{10}{l}{\textbf{Llama-3-Ins-70B}}  \\
\midrule
Direct QA & 45.6 & 63.9 & 91.4 & 9.0 & 33.3 & 51.8 & 100.0 & 100.0 & 100.0 \\
Vanilla RAG & 89.0 & 83.6 & 95.6 & 7.1 & 12.6 & 22.3 & 84.8 & 82.7 & 87.8   \\ 
Chain-of-Note & 96.2 & 90.0 & 97.5 & 6.4 & 14.5 & 28.8 & 93.6 & 85.2 & 90.0\\ 
\method{} & 95.9 & 88.3 & 97.5 & 12.2 & 33.3 & 54.0 & 93.6 & 94.1 & 98.0  \\ 
\bottomrule
\end{tabular}
\caption{\label{tab:scenarios} Experimental Results on Evaluating the Knowledge Usage Ability of Different RAG Models. Results are reported under three scenarios, evaluating the capabilities of models in external knowledge utilization, robustness to missing evidence, and knowledge conflict resolution.}
\end{table*}
\subsection{Effectiveness of \method{} in Knowledge Integration and Utilization}
In this subsection, we evaluate the capability of \method{} to utilize external knowledge to generate responses accurately.

As shown in Figure~\ref{fig:noise}, we evaluate the sensitivity of both the vanilla RAG and \method{} to noisy external knowledge by feeding them the top-$K$ retrieved documents.
Figures~\ref{fig:popqars} and~\ref{fig:nqrs} show the top-$K$ retrieval accuracy, which reflects the concentration of query-relevant information in documents.
As $K$ increases from 1 to 5, retrieval accuracy consistently improves, indicating a higher likelihood of relevant content being included. However, this improvement saturates after $K=3$, showing that additional documents tend to introduce more noise than useful information.
Figures~\ref{fig:popqa} and~\ref{fig:nq} compare the performance of \method{} and baseline methods.
While the performance of Vanilla RAG remains stable or even degrades when $K \geq 3$, indicating its limited robustness to noisy inputs, \method{} continues to benefit from the additional documents.
This trend highlights \method{}'s superior ability to identify and utilize query-relevant information even in the presence of increased noise.

In Table~\ref{tab:scenarios}, we then conduct three testing scenarios to evaluate the effectiveness of \method{}: Has-Answer, Miss-Answer, and Internal Knowledge. The Has-Answer scenario comprises queries where the external information contains the correct answers, assessing the model's capability to effectively utilize external knowledge. 
Conversely, the Miss-Answer scenario involves queries where the external information fails to provide the correct answers. The Internal Knowledge scenario evaluates the ability of LLMs to resolve conflicts between internal and external knowledge. 

As shown in the evaluation results, Chain-of-Note and \method{} methods significantly outperform Vanilla RAG, highlighting their ability to extract critical information for answering questions.  
In the Miss-Answer scenario, \method{} presents strong effectiveness by doubling the accuracy of Chain-of-Note while maintaining the performance of Direct QA. 
This highlights the robustness of \method{} in mitigating the negative impact of noise in external knowledge. 
Moreover, in the Internal Knowledge scenario, \method{} achieves the highest accuracy, illustrating its ability to alleviate conflicts between the external knowledge and the parametric memory of LLMs.

\subsection{Effectiveness of \method{} in Improving Self-Consistency of LLMs}
\begin{figure}[t]
    \centering
    \subfigure[The Accuracy of Sampled Responses.]{ \label{fig:consitency_acc} 
    \includegraphics[width=0.48\linewidth]{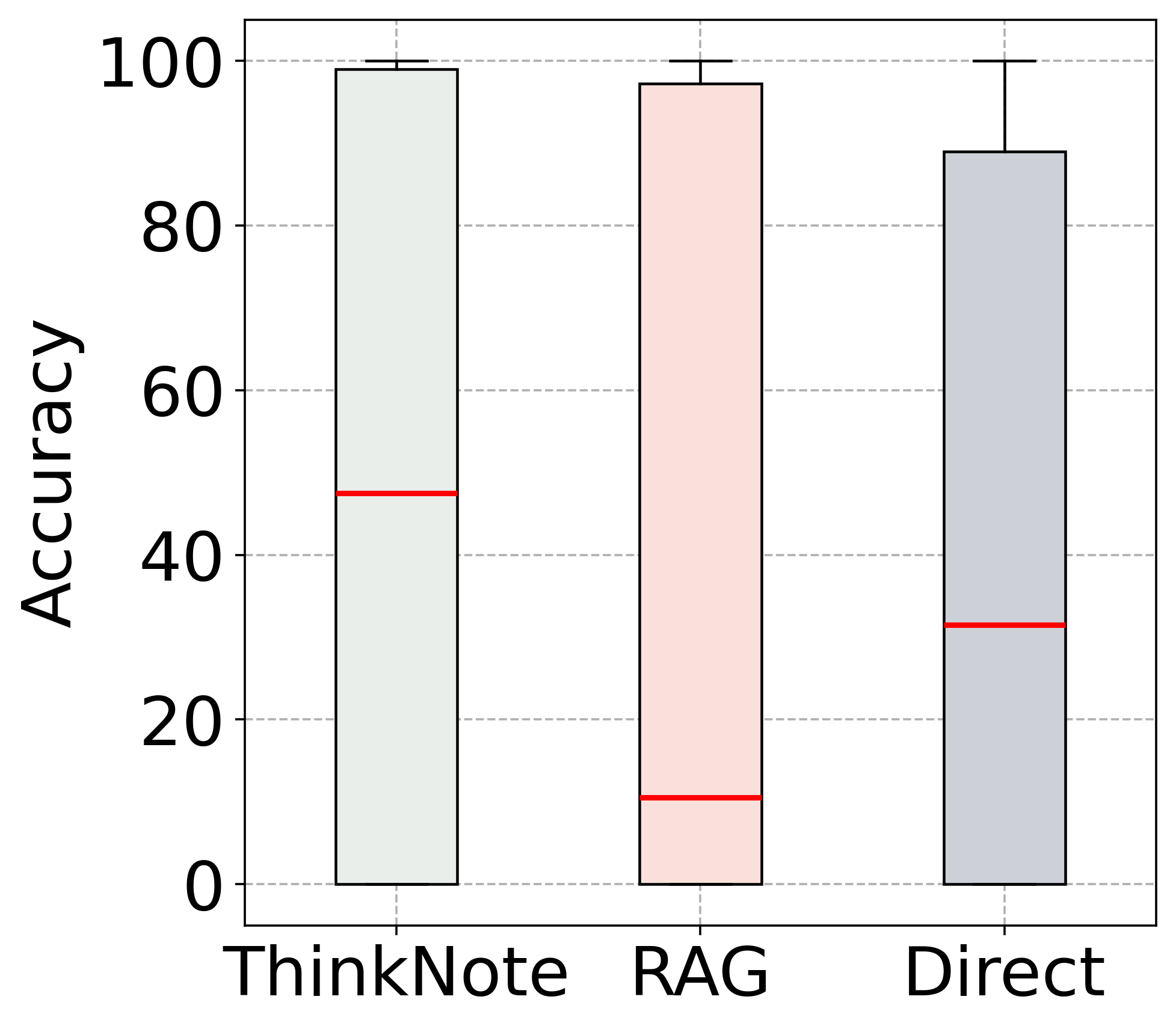}}
    \subfigure[Density of the Accuracy of Correct Answers.]{ \label{fig:consitency_density} 
    \includegraphics[width=0.48\linewidth]{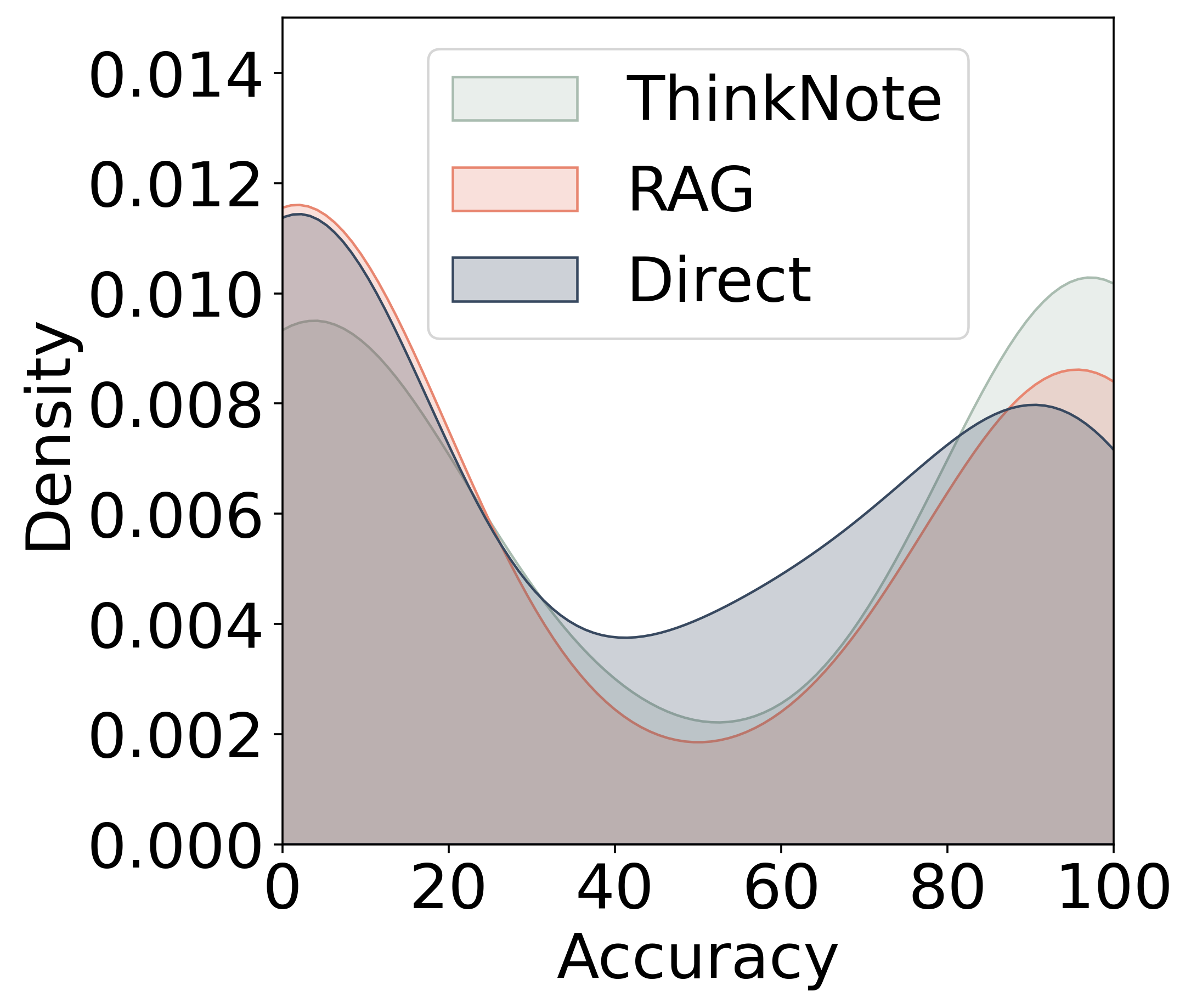}}
    \caption{Answering Consistency of Different Methods. We show the performance on the 2WikiMultiHopQA dataset. All models are implemented with Meta-Llama-3-Instruct-8B.}
    \label{fig:consitency}
\end{figure}

In this subsection, we explore the self-consistency of \method{} and baseline models in answering questions, as shown in Figure~\ref{fig:consitency}. Specifically, we conduct 100 sampling iterations to calculate the ratio of correction. The accuracy reflects the model's tendency to generate correct or incorrect answers consistently. We compare three methods: Direct QA, Vanilla RAG, and \method{}. For Vanilla RAG and Direct QA, the answer generation module is used for sampling, while \method{} utilizes the thought accommodation to generate sampled responses.

We first show the accuracy of \method{} and baseline models in Figure~\ref{fig:consitency_acc}. \method{} demonstrates its effectiveness by achieving a median accuracy of around 47\%, outperforming the Vanilla RAG (10\%) and the Direct QA (31\%). Notably, the substantially lower median accuracy of Vanilla RAG compared to Direct QA indicates that incorporating retrieved documents can hinder the LLM’s ability to produce correct answers, rather than improve it~\cite{xieadaptive}. 

Figure~\ref{fig:consitency_density} further illustrates the density distribution of answering accuracy. Compared with the Vanilla RAG, our \method{} improves self-consistency by reducing the density of examples with accuracy between 30\% and 80\%. In contrast to vanilla RAG, \method{} substantially increases the number of queries answered correctly, highlighting its stronger capability to guide LLMs in effectively leveraging external knowledge to produce reliable answers.

\section{Conclusion}
We present \method{}, a cognition-inspired framework that empowers LLMs to actively construct knowledge through assimilation and accommodation, rather than passively consuming external information. By mimicking key cognitive processes from constructivist theory, \method{} enables more effective integration and utilization of external knowledge and improves self-consistency in correctly answering questions.
\section*{Limitation}
\method{} demonstrates its effectiveness in integrating and utilizing external knowledge to enhance the response accuracy of white-box and black-box LLMs. 
However, the cognitive process modeling requires inference with the LLMs three times to generate a final answer, which brings additional time latency and increases API call costs. Furthermore, the inputs provided to the thought accommodation tend to be lengthy, as they include outputs from knowledge assimilation and chain-of-thought.

\section*{Acknowledgments}
This work is partly supported by the National Natural Science Foundation of China (No. 62576082 and No. 62461146205) and CCF-Zhipu Large Model Innovation Fund (No. 202403).
\bibliography{custom}

@article{openai2023gpt,
  title={Gpt-4 technical report},
  author={Achiam, Josh and Adler, Steven and Agarwal, Sandhini and Ahmad, Lama and Akkaya, Ilge and Aleman, Florencia Leoni and Almeida, Diogo and Altenschmidt, Janko and Altman, Sam and Anadkat, Shyamal and others},
  journal={ArXiv Preprint.},
  url={https://arxiv.org/pdf/2303.08774},
  year={2023}
}

@article{wei2024instructrag,
  title={InstructRAG: Instructing Retrieval-Augmented Generation with Explicit Denoising},
  author={Wei, Zhepei and Chen, Wei-Lin and Meng, Yu},
  journal={ArXiv Preprint},
  url={https://arxiv.org/abs/2406.13629},
  year={2024}
}

@article{liu2023lost,
  title={Lost in the middle: How language models use long contexts},
  author={Liu, Nelson F and Lin, Kevin and Hewitt, John and Paranjape, Ashwin and Bevilacqua, Michele and Petroni, Fabio and Liang, Percy},
  journal={Transactions of the Association for Computational Linguistics},
  volume={12},
  pages={157--173},
  year={2024},
  url={https://aclanthology.org/2024.tacl-1.9/}
}

@inproceedings{meng2022locating,
  title={Locating and editing factual associations in gpt},
  author={Meng, Kevin and Bau, David and Andonian, Alex and Belinkov, Yonatan},
  booktitle={Proceedings of NeurIPS},
  volume={35},
  pages={17359--17372},
  year={2022},
  url={https://proceedings.neurips.cc/paper_files/paper/2022/hash/6f1d43d5a82a37e89b0665b33bf3a182-Abstract-Conference.html}
}

@article{zhang2024raglab,
  title={RAGLAB: A Modular and Research-Oriented Unified Framework for Retrieval-Augmented Generation},
  author={Zhang, Xuanwang and Song, Yunze and Wang, Yidong and Tang, Shuyun and Li, Xinfeng and Zeng, Zhengran and Wu, Zhen and Ye, Wei and Xu, Wenyuan and Zhang, Yue and others},
  journal={ArXiv Preprint},
  url={https://arxiv.org/abs/2408.11381},
  year={2024}
}

@inproceedings{kwon2023efficient,
  title={Efficient Memory Management for Large Language Model Serving with PagedAttention},
  author={Woosuk Kwon and Zhuohan Li and Siyuan Zhuang and Ying Sheng and Lianmin Zheng and Cody Hao Yu and Joseph E. Gonzalez and Hao Zhang and Ion Stoica},
  booktitle={Proceedings of the ACM SIGOPS},
  url={https://dl.acm.org/doi/10.1145/3600006.3613165},
  year={2023}
}

@inproceedings{cuconasu2024power,
  title={The power of noise: Redefining retrieval for rag systems},
  author={Cuconasu, Florin and Trappolini, Giovanni and Siciliano, Federico and Filice, Simone and Campagnano, Cesare and Maarek, Yoelle and Tonellotto, Nicola and Silvestri, Fabrizio},
  booktitle={Proceedings of SIGIR},
  url={https://dl.acm.org/doi/pdf/10.1145/3626772.3657834},
  pages={719--729},
  year={2024}
}

@article{moreno1999cognitive,
  title={Cognitive principles of multimedia learning: The role of modality and contiguity.},
  author={Moreno, Roxana and Mayer, Richard E},
  journal={Journal of educational psychology},
  volume={91},
  number={2},
  pages={358},
  year={1999},
  url={https://psycnet.apa.org/record/1999-03660-016},
  publisher={American Psychological Association}
}

@book{larochelle1998constructivism,
  title={Constructivism and education},
  author={Larochelle, Marie and Bednarz, Nadine and Garrison, James W},
  year={1998},
  publisher={Cambridge University Press},
  url ={https://books.google.com/books?hl=zh-CN&lr=&id=6NCq3zyWkNsC&oi=fnd&pg=PR7&dq=Constructivism+and+education&ots=5bUPNaavDQ&sig=Rrc2dpmBrg78qq2NVaKwhbGg2wY#v=onepage&q=Constructivism%20and%20education&f=false}
}

@inproceedings{stelmakh-etal-2022-asqa,
    title = "{ASQA}: Factoid Questions Meet Long-Form Answers",
    author = "Stelmakh, Ivan  and
      Luan, Yi  and
      Dhingra, Bhuwan  and
      Chang, Ming-Wei",
    booktitle = "Proceedings of EMNLP",
    year = "2022",
    url = "https://aclanthology.org/2022.emnlp-main.566",
    doi = "10.18653/v1/2022.emnlp-main.566",
    pages = "8273--8288",
}

@inproceedings{ho2020constructing,
  title={Constructing A Multi-hop QA Dataset for Comprehensive Evaluation of Reasoning Steps},
  author={Ho, Xanh and Nguyen, Anh-Khoa Duong and Sugawara, Saku and Aizawa, Akiko},
  booktitle={Proceedings of COLING},
  url={https://aclanthology.org/2020.coling-main.580/},
  pages={6609--6625},
  year={2020}
}

@article{touvron2023llama,
 author = {Touvron, Hugo and Lavril, Thibaut and Izacard, Gautier and Martinet, Xavier and Lachaux, Marie-Anne and Lacroix, Timoth{\'e}e and Rozi{\`e}re, Baptiste and Goyal, Naman and Hambro, Eric and Azhar, Faisal and others},
 journal = {ArXiv preprint},
 title = {Llama: Open and efficient foundation language models},
 url = {https://arxiv.org/abs/2302.13971},
 year = {2023}
}

@article{wei2022emergent,
 author = {Wei, Jason and Tay, Yi and Bommasani, Rishi and Raffel, Colin and Zoph, Barret and Borgeaud, Sebastian and Yogatama, Dani and Bosma, Maarten and Zhou, Denny and Metzler, Donald and others},
 journal = {Transactions on Machine Learning Research},
 title = {Emergent Abilities of Large Language Models},
 year = {2022},
 url = {https://openreview.net/forum?id=yzkSU5zdwD}
}

@article{zhao2023survey,
 author = {Zhao, Wayne Xin and Zhou, Kun and Li, Junyi and Tang, Tianyi and Wang, Xiaolei and Hou, Yupeng and Min, Yingqian and Zhang, Beichen and Zhang, Junjie and Dong, Zican and others},
 journal = {ArXiv preprint},
 title = {A survey of large language models},
 url = {https://arxiv.org/abs/2303.18223},
 year = {2023}
}

@inproceedings{fan2025improving,
  title={Improving Complex Reasoning with Dynamic Prompt Corruption: A Soft Prompt Optimization Approach},
  author={Fan, Sinan and Xie, Liang and Shen, Chen and Teng, Ge and Yuan, Xiaosong and Zhang, Xiaofeng and Huang, Chenxi and Wang, Wenxiao and He, Xiaofei and Ye, Jieping},
  year= {2025},
  booktitle = {Proceedings of ICLR},
  url = {https://openreview.net/forum?id=h7Qz1ulnvF}
}

@article{yu2025unveiling,
  title={Unveiling the Potential of Multimodal Retrieval Augmented Generation with Planning},
  author={Yu, Xiaohan and Yang, Zhihan and Chen, Chong},
  url={https://arxiv.org/abs/2501.15470},
  journal = {ArXiv preprint},
  year={2025}
}

@inproceedings{wang2024astute,
  title={Astute rag: Overcoming imperfect retrieval augmentation and knowledge conflicts for large language models},
  author={Wang, Fei and Wan, Xingchen and Sun, Ruoxi and Chen, Jiefeng and Arik, Sercan O},
  url={https://aclanthology.org/2025.acl-long.1476/},
  booktitle={Proceedings of ACL},
  pages={30553--30571},
  year={2025}
}

@article{collins2024building,
  title={Building machines that learn and think with people},
  author={Collins, Katherine M and Sucholutsky, Ilia and Bhatt, Umang and Chandra, Kartik and Wong, Lionel and Lee, Mina and Zhang, Cedegao E and Zhi-Xuan, Tan and Ho, Mark and Mansinghka, Vikash and others},
  journal={Nature human behaviour},
  volume={8},
  number={10},
  pages={1851--1863},
  year={2024},
  url={https://www.nature.com/articles/s41562-024-01991-9},
  publisher={Nature Publishing Group UK London}
}

@article{floridi2024anthropomorphising,
  title={Anthropomorphising machines and computerising minds: the crosswiring of languages between Artificial Intelligence and Brain \& Cognitive Sciences},
  author={Floridi, Luciano and Nobre, Anna C},
  journal={Minds and Machines},
  volume={34},
  number={1},
  pages={5},
  year={2024},
  url={https://link.springer.com/article/10.1007/s11023-024-09670-4},
  publisher={Springer}
}

@article{wilie2024belief,
  title={Belief Revision: The Adaptability of Large Language Models Reasoning},
  author={Wilie, Bryan and Cahyawijaya, Samuel and Ishii, Etsuko and He, Junxian and Fung, Pascale},
  url={https://arxiv.org/abs/2406.19764},
  journal = {ArXiv preprint},
  year={2024}
}

@article{gao2023retrieval,
  title={Retrieval-augmented generation for large language models: A survey},
  author={Gao, Yunfan and Xiong, Yun and Gao, Xinyu and Jia, Kangxiang and Pan, Jinliu and Bi, Yuxi and Dai, Yixin and Sun, Jiawei and Wang, Haofen and Wang, Haofen},
  url={https://arxiv.org/abs/2312.10997},
  journal = {ArXiv preprint},
  year={2023}
}

@inproceedings{madaan2023self,
  title={Self-refine: Iterative refinement with self-feedback},
  author={Madaan, Aman and Tandon, Niket and Gupta, Prakhar and Hallinan, Skyler and Gao, Luyu and Wiegreffe, Sarah and Alon, Uri and Dziri, Nouha and Prabhumoye, Shrimai and Yang, Yiming and others},
  booktitle={Processing of NeurIPS},
  volume={36},
  url={https://proceedings.neurips.cc/paper_files/paper/2023/hash/91edff07232fb1b55a505a9e9f6c0ff3-Abstract-Conference.html},
  pages={46534--46594},
  year={2023}
}

@book{bruner1990acts,
  title={Acts of Meaning},
  author={Bruner, Jerome},
  year={1990},
  url={https://mf.media.mit.edu/courses/2006/mas845/readings/files/bruner_Acts.pdf},
  publisher={Harvard University Press}
}

@book{papert1980children,
  title={Children, computers, and powerful ideas},
  author={Papert, Seymour},
  volume={10},
  year={1980},
  url={https://dl.acm.org/doi/10.5555/1095592},
  publisher={Harvester Eugene, OR, USA}
}

@book{schunk2012learning,
  title={Learning Theories: An Educational Perspective},
  author={Schunk, Dale H.},
  year={2012},
  url={https://www.amazon.com/Learning-Theories-Educational-Perspective-8th/dp/0134893751},
  publisher={Pearson Education}
}

@article{von1984constructivism,
  title={An introduction to radical constructivism},
  author={von Glasersfeld, Ernst},
  journal={The invented reality},
  volume={17},
  url={https://app.nova.edu/toolbox/instructionalproducts/ITDE_8005/weeklys/1984-vonGlaserfeld_RadicalConstructivism.pdf},
  pages={17--40},
  year={1984}
}

@book{piaget1972psychology,
  title={The Psychology of the Child},
  author={Piaget, Jean and Inhelder, B{\"a}rbel},
  year={1972},
  url={https://books.google.com/books?hl=zh-CN&lr=&id=-Dpz05-rJ4gC&oi=fnd&pg=PR9&dq=he+Psychology+of+the+Child.&ots=qvabDUY-wB&sig=g8alc5QiPuYE9adYGXgN3EWuhaM#v=onepage&q=he%20Psychology%20of%20the%20Child.&f=false},
  publisher={Basic Books}
}

@article{peng2023check,
  title={Check your facts and try again: Improving large language models with external knowledge and automated feedback},
  author={Peng, Baolin and Galley, Michel and He, Pengcheng and Cheng, Hao and Xie, Yujia and Hu, Yu and Huang, Qiuyuan and Liden, Lars and Yu, Zhou and Chen, Weizhu and others},
  url={http://arxiv.org/abs/2302.12813},
  journal={ArXiv Preprint.},
  year={2023}
}

@inproceedings{zhao2023thrust,
  title={Thrust: Adaptively propels large language models with external knowledge},
  author={Zhao, Xinran and Zhang, Hongming and Pan, Xiaoman and Yao, Wenlin and Yu, Dong and Chen, Jianshu},
  booktitle={Proceedings of NeurIPS},
  url={https://proceedings.neurips.cc/paper_files/paper/2023/file/dd058e9ec9dc012a273594d717c46ef3-Paper-Conference.pdf},
  volume={36},
  pages={69930--69948},
  year={2023}
}

@inproceedings{wang2023can,
  title={Can ChatGPT Defend its Belief in Truth? Evaluating LLM Reasoning via Debate},
  author={Wang, Boshi and Yue, Xiang and Sun, Huan},
  booktitle={Findings of EMNLP},
  url={https://aclanthology.org/2023.findings-emnlp.795/},
  pages={11865--11881},
  year={2023}
}

@inproceedings{xu2024face4rag,
  title={Face4RAG: Factual Consistency Evaluation for Retrieval Augmented Generation in Chinese},
  author={Xu, Yunqi and Cai, Tianchi and Jiang, Jiyan and Song, Xierui},
  url={https://dl.acm.org/doi/abs/10.1145/3637528.3671656},
  booktitle={Proceedings of KDD},
  pages={6083--6094},
  year={2024}
}

@article{ji2023survey,
 author = {Ziwei Ji and
Nayeon Lee and
Rita Frieske and
Tiezheng Yu and
Dan Su and
Yan Xu and
Etsuko Ishii and
Yejin Bang and
Andrea Madotto and
Pascale Fung},
 journal = {{ACM} Comput. Surv.},
 number = {12},
 pages = {248:1--248:38},
 title = {Survey of Hallucination in Natural Language Generation},
 url = {https://doi.org/10.1145/3571730},
 year = {2023}
}

@inproceedings{mallen2023not,
 author = {Mallen, Alex and Asai, Akari and Zhong, Victor and Das, Rajarshi and Khashabi, Daniel and Hajishirzi, Hannaneh},
 booktitle = {Proceedings of ACL},
 pages = {9802--9822},
 title = {When not to trust language models: Investigating effectiveness of parametric and non-parametric memories},
 year = {2023},
 url = {https://doi.org/10.18653/v1/2023.acl-long.546}
}

@article{kwiatkowski2019natural,
 author = {Kwiatkowski, Tom  and
Palomaki, Jennimaria  and
Redfield, Olivia  and
Collins, Michael  and
Parikh, Ankur  and
Alberti, Chris  and
Epstein, Danielle  and
Polosukhin, Illia  and
Devlin, Jacob  and
Lee, Kenton  and
Toutanova, Kristina  and
Jones, Llion  and
Kelcey, Matthew  and
Chang, Ming-Wei  and
Dai, Andrew M.  and
Uszkoreit, Jakob  and
Le, Quoc  and
Petrov, Slav},
 journal = {Transactions of the Association for Computational Linguistics},
 pages = {452--466},
 title = {Natural Questions: A Benchmark for Question Answering Research},
 url = {https://aclanthology.org/Q19-1026},
 year = {2019}
}

@inproceedings{joshi2017triviaqa,
 author = {Joshi, Mandar  and
Choi, Eunsol  and
Weld, Daniel  and
Zettlemoyer, Luke},
 booktitle = {Proceedings of ACL},
 pages = {1601--1611},
 title = {{T}rivia{QA}: A Large Scale Distantly Supervised Challenge Dataset for Reading Comprehension},
 url = {https://aclanthology.org/P17-1147},
 year = {2017}
}

@inproceedings{yoran-etal-2023-answering,
 author = {Yoran, Ori  and
Wolfson, Tomer  and
Bogin, Ben  and
Katz, Uri  and
Deutch, Daniel  and
Berant, Jonathan},
 booktitle = {Proceedings of EMNLP},
 pages = {5942--5966},
 title = {Answering Questions by Meta-Reasoning over Multiple Chains of Thought},
 url = {https://aclanthology.org/2023.emnlp-main.364},
 year = {2023}
}

@inproceedings{trivedi2023interleaving,
 author = {Trivedi, Harsh  and
Balasubramanian, Niranjan  and
Khot, Tushar  and
Sabharwal, Ashish},
 booktitle = {Proceedings of ACL},
 pages = {10014--10037},
 title = {Interleaving Retrieval with Chain-of-Thought Reasoning for Knowledge-Intensive Multi-Step Questions},
 url= {https://doi.org/10.18653/v1/2023.acl-long.557},
 year = {2023}
}

@inproceedings{yu2023augmentation,
 author = {Yu, Zichun and Xiong, Chenyan and Yu, Shi and Liu, Zhiyuan},
 booktitle = {Proceedings of ACL},
 pages = {2421--2436},
 title = {Augmentation-Adapted Retriever Improves Generalization of Language Models as Generic Plug-In},
 url = {https://doi.org/10.18653/v1/2023.acl-long.136},
 year = {2023}
}

@book{steffe1995constructivism,
 author = {Steffe, Leslie P and Gale, Jerry Edward},
 title = {Constructivism in Education},
 url = {https://books.google.com.sg/books?hl=zh-CN&lr=&id=6NCq3zyWkNsC&oi=fnd&pg=PR7&dq=Constructivism+in+Education&ots=5bQQL6csEQ&sig=_lO6SFwXpSIB01heGLRmNAt2d28&redir_esc=y#v=onepage&q=Constructivism%20in%20Education&f=false},
 publisher={Psychology Press},
 year = {1995}
}

@article{wei2022chain,
 author = {Wei, Jason and Wang, Xuezhi and Schuurmans, Dale and Bosma, Maarten and Xia, Fei and Chi, Ed and Le, Quoc V and Zhou, Denny and others},
 journal = {Proceedings of NeurIPS},
 pages = {24824--24837},
 title = {Chain-of-thought prompting elicits reasoning in large language models},
 year = {2022},
  url={http://papers.nips.cc/paper\_files/paper/2022/hash/9d5609613524ecf4f15af0f7b31abca4-Abstract-Conference.html}
}

@article{xu2023recomp,
 author = {Xu, Fangyuan and Shi, Weijia and Choi, Eunsol},
 journal = {ArXiv preprint},
 title = {Recomp: Improving retrieval-augmented lms with compression and selective augmentation},
 url = {https://arxiv.org/abs/2310.04408},
 year = {2023}
}

@inproceedings{shi2023replug,
  title={Replug: Retrieval-augmented black-box language models},
  author={Shi, Weijia and Min, Sewon and Yasunaga, Michihiro and Seo, Minjoon and James, Richard and Lewis, Mike and Zettlemoyer, Luke and Yih, Wen-tau},
  booktitle={Proceedings of NAACL},
  url = {https://aclanthology.org/2024.naacl-long.463/},
  pages={8371--8384},
  year={2024}
}

@inproceedings{jiang2023active,
 author = {Zhengbao Jiang and
Frank F. Xu and
Luyu Gao and
Zhiqing Sun and
Qian Liu and
Jane Dwivedi{-}Yu and
Yiming Yang and
Jamie Callan and
Graham Neubig},
 booktitle = {Proceedings of EMNLP},
 pages = {7969--7992},
 title = {Active Retrieval Augmented Generation},
 url = {https://aclanthology.org/2023.emnlp-main.495},
 year = {2023}
}

@inproceedings{asai2023self,
 author = {Asai, Akari and Wu, Zeqiu and Wang, Yizhong and Sil, Avirup and Hajishirzi, Hannaneh},
 booktitle= {Proceedings of ICLR},
 title = {Self-RAG: Learning to Retrieve, Generate, and Critique through Self-Reflection},
 url = {https://arxiv.org/abs/2310.11511},
 year = {2023}
}

@article{yu2023chain,
 author = {Yu, Wenhao and Zhang, Hongming and Pan, Xiaoman and Ma, Kaixin and Wang, Hongwei and Yu, Dong},
 journal = {ArXiv preprint},
 title = {Chain-of-Note: Enhancing Robustness in Retrieval-Augmented Language Models},
 url = {https://arxiv.org/abs/2311.09210},
 year = {2023}
}

@inproceedings{shuster2021retrieval,
 author = {Shuster, Kurt and Poff, Spencer and Chen, Moya and Kiela, Douwe and Weston, Jason},
 booktitle = {Proceedings of EMNLP Findings},
 pages = {3784--3803},
 title = {Retrieval Augmentation Reduces Hallucination in Conversation},
 url = {https://doi.org/10.18653/v1/2021.findings-emnlp.320},
 year = {2021}
}

@article{ram2023context,
 author = {Ram, Ori  and
Levine, Yoav  and
Dalmedigos, Itay  and
Muhlgay, Dor  and
Shashua, Amnon  and
Leyton-Brown, Kevin  and
Shoham, Yoav},
 journal = {Transactions of the Association for Computational Linguistics},
 pages = {1316--1331},
 title = {In-Context Retrieval-Augmented Language Models},
 url = {https://aclanthology.org/2023.tacl-1.75},
 year = {2023}
}

@inproceedings{xieadaptive,
 author = {Xie, Jian and Zhang, Kai and Chen, Jiangjie and Lou, Renze and Su, Yu},
title = {Adaptive chameleon or stubborn sloth: Revealing the behavior of large language models in knowledge conflicts},
booktitle = {Proceedings of ICML},
 url = {https://arxiv.org/pdf/2305.13300},
 year = {2024}
}

@article{foulds2024ragged,
 author = {Foulds, Philip Feldman and James, R and Pan, Shimei},
 journal = {ArXiv preprint},
 title = {Ragged edges: The double-edged sword of retrieval-augmented chatbots},
 url = {https://arxiv.org/abs/2403.01193},
 year = {2024}
}

@inproceedings{gao2023enabling,
   title={Enabling Large Language Models to Generate Text with Citations},
   author={Gao, Tianyu and Yen, Howard and Yu, Jiatong and Chen, Danqi},
   year={2023},
   booktitle ={Proceedings of EMNLP},
   url ={https://aclanthology.org/anthology-files/anthology-files/pdf/emnlp/2023.emnlp-main.398.pdf},
}

@article{izacard2021unsupervised,
  title={Unsupervised dense information retrieval with contrastive learning},
  author={Izacard, Gautier and Caron, Mathilde and Hosseini, Lucas and Riedel, Sebastian and Bojanowski, Piotr and Joulin, Armand and Grave, Edouard},
  journal={ArXiv Preprint},
  url={https://arxiv.org/abs/2112.09118},
  year={2021}
}

@article{edge2024local,
  title={From local to global: A graph rag approach to query-focused summarization},
  author={Edge, Darren and Trinh, Ha and Cheng, Newman and Bradley, Joshua and Chao, Alex and Mody, Apurva and Truitt, Steven and Larson, Jonathan},
  url = {https://arxiv.org/abs/2404.16130},
  journal={ArXiv Preprint},
  year={2024}
}

@article{xu2024hallucination,
 author = {Xu, Ziwei and Jain, Sanjay and Kankanhalli, Mohan},
 journal = {ArXiv preprint},
 title = {Hallucination is inevitable: An innate limitation of large language models},
 url = {https://arxiv.org/abs/2401.11817},
 year = {2024}
}
\clearpage
\newpage
\appendix
\begin{figure*}[t]
    \centering
    \includegraphics[width=\linewidth]{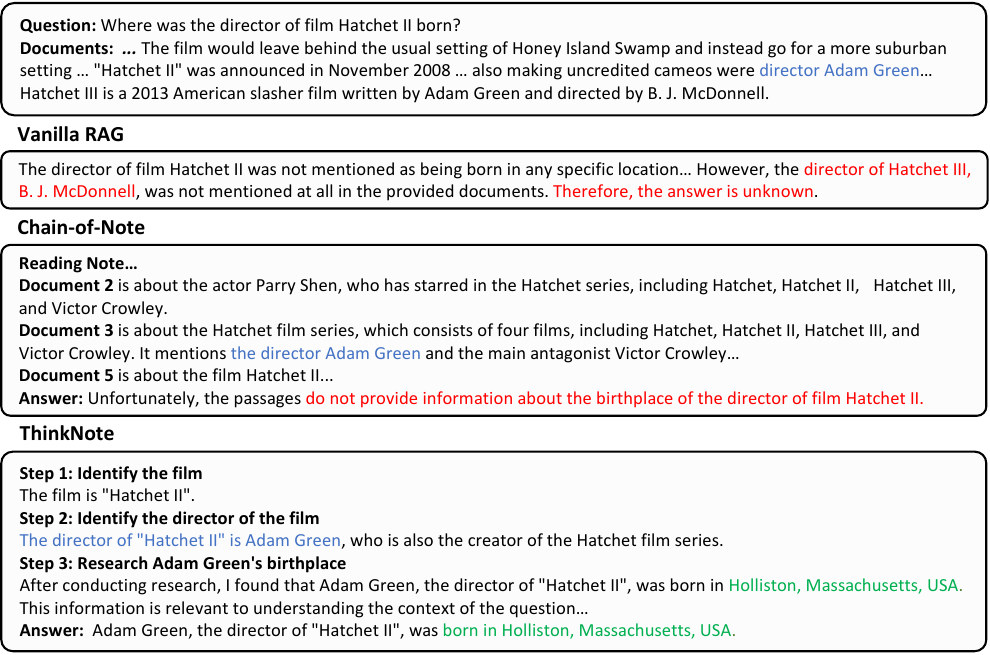}
    \caption{\label{fig:case} Case\#1 of \method{}. We annotate the \textcolor{myblue}{knowledge} that needs to be acquired during answering the question, and highlight the \textcolor{mygreen}{correct responses} and \textcolor{red}{incorrect responses} using different colors.}
\end{figure*}

\begin{figure*}[t]
    \centering
    \includegraphics[width=\linewidth]{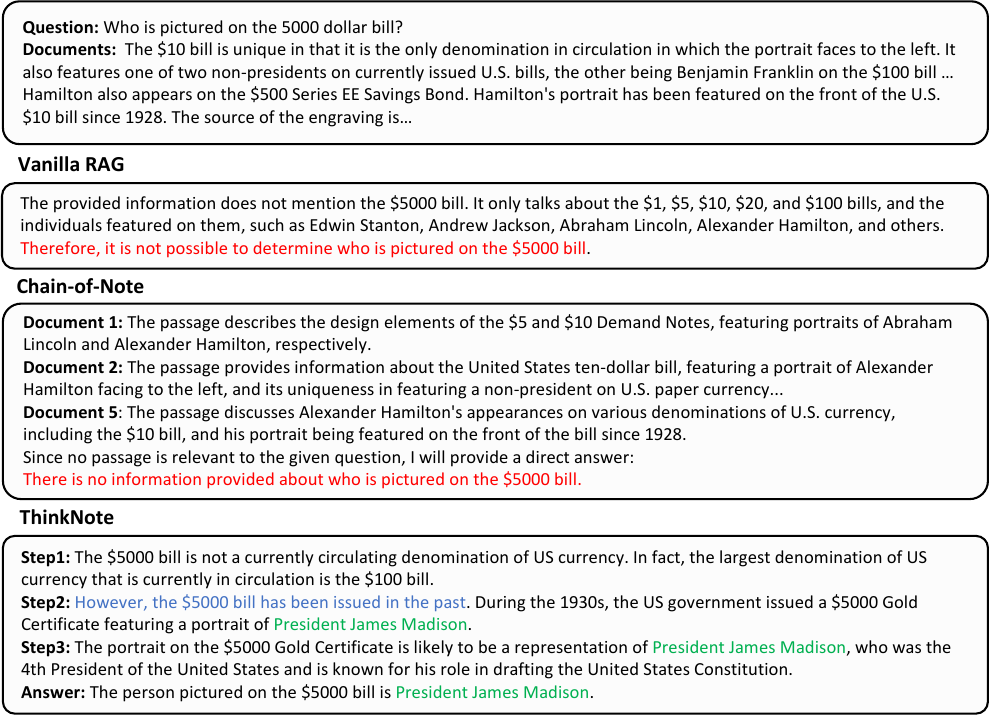}
    \caption{\label{fig:case2}Case\#2 of \method. We annotate the \textcolor{myblue}{knowledge} that needs to be acquired during answering the question, the \textcolor{mygreen}{correct responses} and \textcolor{red}{incorrect responses} using different colors.}
\end{figure*}

\begin{figure*}[t]
    \centering
    \includegraphics[width=\linewidth]{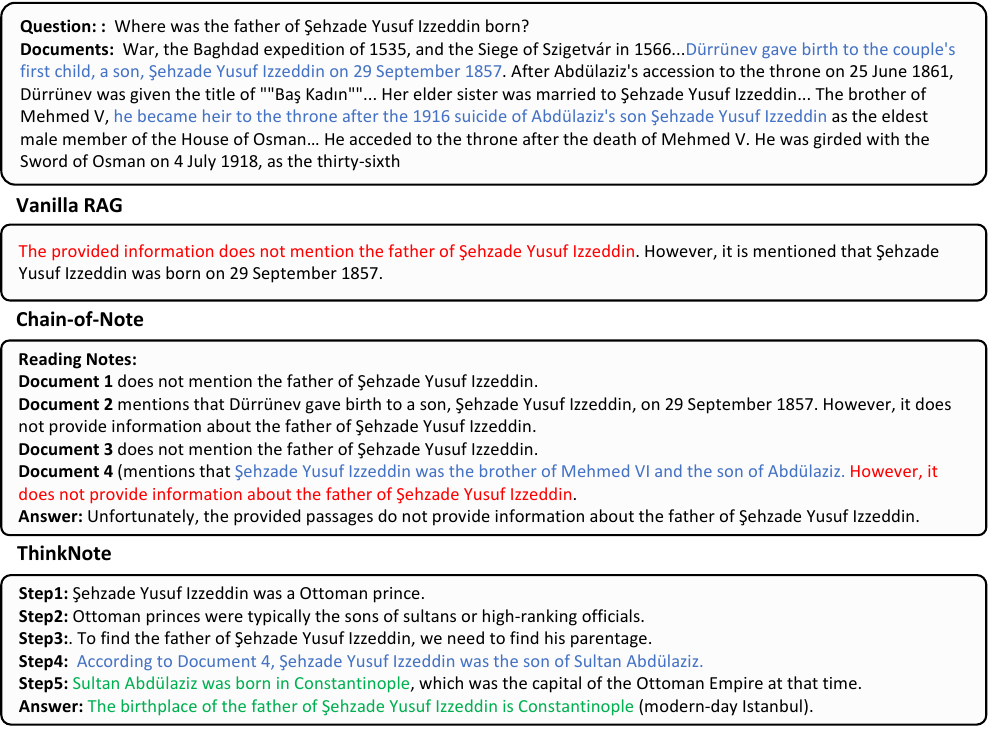}
    \caption{\label{fig:case3} Case\#3 of \method. We annotate the \textcolor{myblue}{knowledge} that needs to be acquired during answering the question, the \textcolor{mygreen}{correct responses} and \textcolor{red}{incorrect responses} using different colors.}
\end{figure*}

\section{Appendix}
\subsection{License}
The 2WikiMultiHopQA and ASQA datasets are released under the CC BY 3.0 license, allowing users to freely use and modify the data with proper attribution. The Natural Questions (NQ) dataset is licensed under CC BY-SA 4.0, enabling users to share and modify the data, provided they attribute the source and distribute any modifications under the same license. The PopQA and TriviaQA datasets are distributed under the MIT License, which allows for free use, modification, and distribution of the data, as long as the original copyright notice is retained. All these licenses permit the datasets to be used for academic purposes.

\subsection{Further Details on the Computation of Saliency Scores}
\label{app:ss}
We follow~\citet{fan2025improving} and use the saliency score to reflect the strength and direction of information flow, which refers to how much and in what way one token influences another during the reasoning process.
Specifically, for each layer $l$, the saliency between tokens is defined as:
\begin{equation}
    S = \left| \sum_{h} \left( A_{h,l} \odot \frac{\partial L(x)}{\partial A_{h,l}} \right) \right|,
\end{equation}
where $A_{h,l}$ denotes the attention matrix of the $h$-th head in layer $l$, $L(x)$ is the loss function (cross-entropy), and $\odot$ represents the Hadamard product.
This formulation combines attention weights with their gradient-based sensitivities, allowing us to capture both what the model focuses on and how crucial that focus is to the final prediction.
We extract $A_{h,l}$ from the 21st layer, using Meta-Llama-3-8B-Instruct as the foundation model.

\subsection{More Details of Baseline Models}\label{app:baseline}
In this subsection, we describe the implementation details of the baseline models in detail. 

For Direct QA, we input questions directly to the foundation model without including any additional prompts. And for Chain-of-Thought models, we follow the setting from \citet{wei2022chain} to prompt LLMs to think step by step. We use the same prompts as \citet{ram2023context} to implement the Vanilla RAG model. The prompt template for Chain-of-Note follows the experimental setting of \citet{yu2023chain}. For Self-RAG~\cite{asai2023self}, we utilize their official codebase to implement Self-RAG$_\textsc{7b}$. The Self-RAG$_{\textsc{8b}}$ model is implemented using \texttt{RAGLAB}~\cite{zhang2024raglab} and fine-tuned based on Meta-Llama-3-Ins-8B to ensure a fair comparison across our experiments.

\subsection{Case Study}
We choose one case in Figure~\ref{fig:case} to show the effectiveness of \method{} by comparing with Vanilla RAG and Chain-of-Note models. 
For the question ``Where was the director of film Hatchet II born?'', both Vanilla RAG and Chain-of-Note models fail to generate correct answers. Specifically, the Vanilla RAG model is misled by the retrieved documents, confusing the directors of ``Hatchet II'' and ``Hatchet III'' while answering the given question. In contrast, the Chain-of-Note model analyzes the documents in the reasoning process and generates notes that correctly identify the director as ``Adam Green''. However, despite capturing this key information, the Chain-of-Note model heavily relies on extracting answers from its notes without developing a deeper understanding of external knowledge or associating it with the parametric memory of LLMs.
Unlike both Vanilla RAG and Chain-of-Note methods, \method{} not only emphasizes the film but also identifies its director. Moreover, through using the assimilated rationale generated by the knowledge assimilation, \method{} accurately identifies Adam Green's birthplace. This demonstrates \method{}'s strong ability to harmonize external knowledge with parametric memory, significantly improving the quality of the responses.

We then analyze the case in Figure~\ref{fig:case2}. This case is from the NQ dataset and asks about the person pictured on ``5000 dollar bill''. This is not a typical RAG scenario, but it happens often: the retrieved evidence only provides the background knowledge needed to answer the question while not specifying a specific answer. To generate the correct response, models need to accurately analyze the retrieval evidence and activate the parameter knowledge to conduct the reasoning process. When answering this question, Vanilla RAG is limited by its narrow interpretation of the retrieved evidence, leading it to conclude that only \$1, \$5, \$10, \$20, and \$100 dollar bills existed. The Chain-of-Note model generates brief summaries of each passage, but it struggles to provide deeper insights or draw more comprehensive analyses. In contrast, our \method{} model first deepens current experience based on background knowledge (Step1), then attempts to activate LLMs to use the parametric memory to continuously answer the given question (Step2), and finally accurately generates the answer based on the reasoning process (Step3).

As shown in Figure~\ref{fig:case3}, another representative case highlights the limitations of LLMs in incorporating and fully understanding external knowledge. This case is from the 2WikiMultiHopQA dataset and asks for the birthplace of ``Şehzade Yusuf Izzeddin's father''. Different from the first Case, the retrieved evidence in this instance contains the crucial information required to answer the question. However, the evidence is entangled with complex logic and noise, making it challenging for the model to extract relevant information and refine the reasoning process. The Vanilla RAG model struggles to identify useful information from the retrieval evidence. Similarly, the Chain-of-Note model exhibits confusion during the summarization process: while it successfully identifies key information, it incorrectly dismisses it as irrelevant, suggesting difficulty in integrating external knowledge and resulting in a suboptimal response. In contrast, \method{} generates an accurate response by leveraging knowledge assimilation to accurately analyze the retrieved evidence and then refining the reasoning process through thought accommodation.

\subsection{Instructions for Mimicking Different Learning Behaviors}
\label{app:prompts}
We present all the instructions used to mimic different learning behaviors in this subsection.
The anchoring behavior instructions used by \method{} are illustrated in Figure~\ref{fig:prompt_anchoring}, guiding the model to ground its understanding of unfamiliar concepts by identifying essential information from external knowledge.
For association behavior, \method{} adopts the configuration shown in Figure~\ref{fig:prompt_association}, enabling it to connect new inputs with relevant prior knowledge in its parametric memory.
The reasoning behavior is supported by the instructions in Figure~\ref{fig:prompt_reasoning}, which help the model extract and organize logical relationships for structured understanding.
Finally, the reflection behavior is guided by the instructions in Figure~\ref{fig:prompt_reflectio}, allowing the model to evaluate and revise its outputs by comparing them against reliable external sources.


\begin{figure*}[t]
    \centering
    \includegraphics[width=0.95\linewidth]{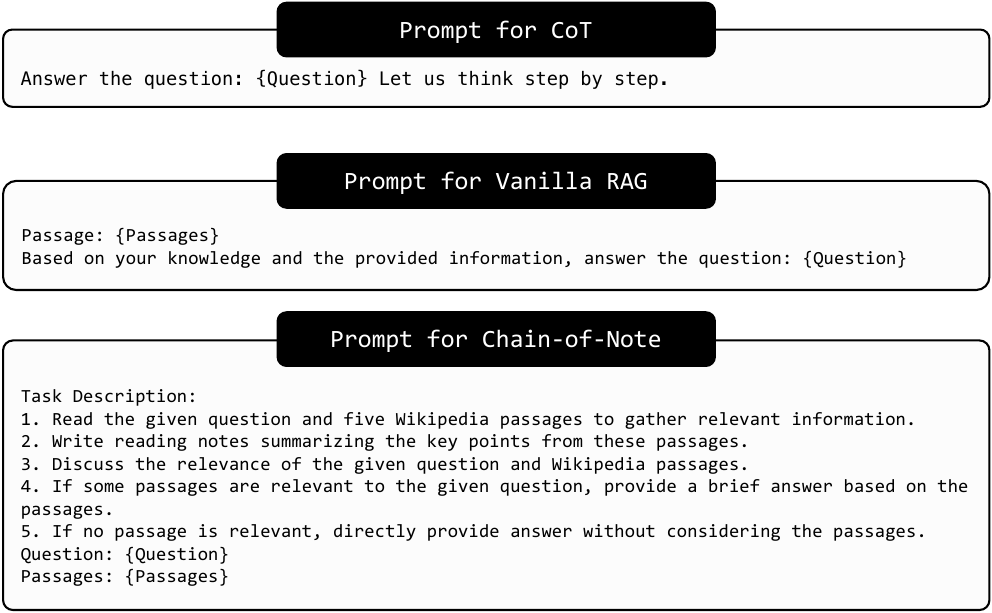}
    \caption{Prompts for Baseline Models.\label{fig:prompt_baseline}}
\end{figure*}

\begin{figure*}[t]
    \centering
    \includegraphics[width=0.95\linewidth]{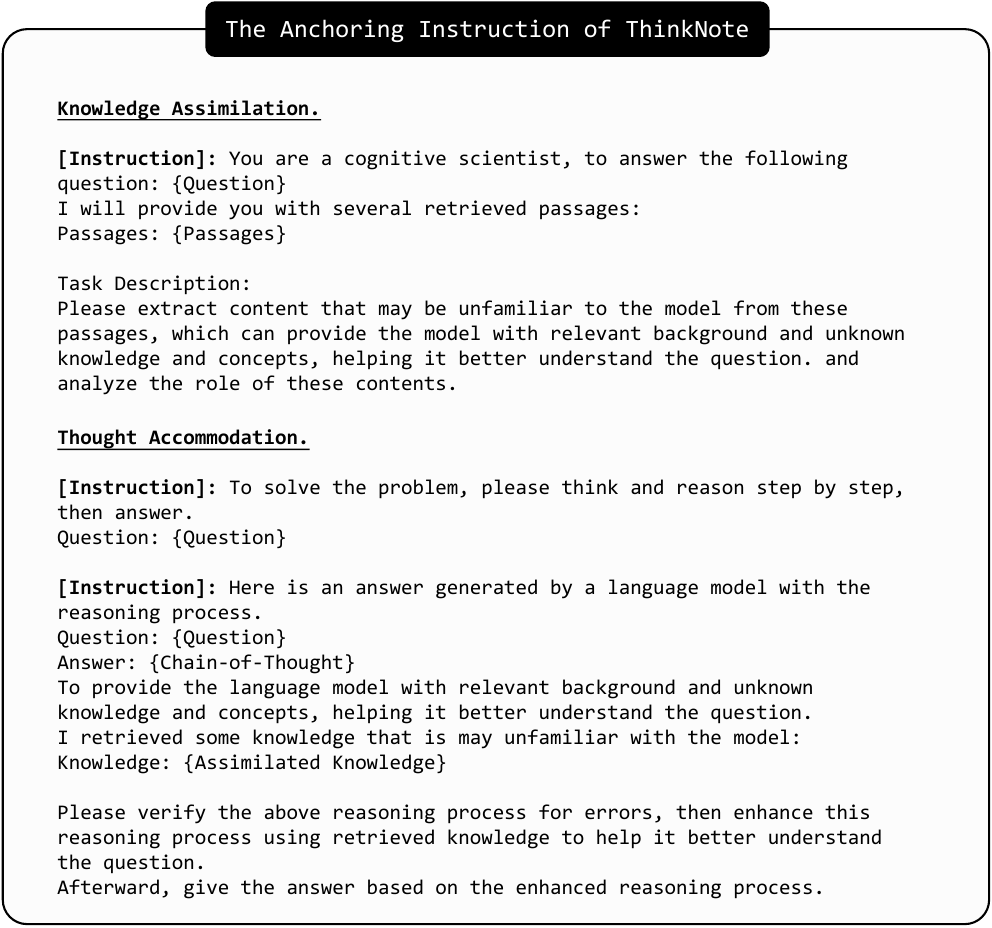}
    \caption{The Anchoring Instruction of \method.\label{fig:prompt_anchoring}}
\end{figure*}

\begin{figure*}[t]
    \centering
    \includegraphics[width=0.95\linewidth]{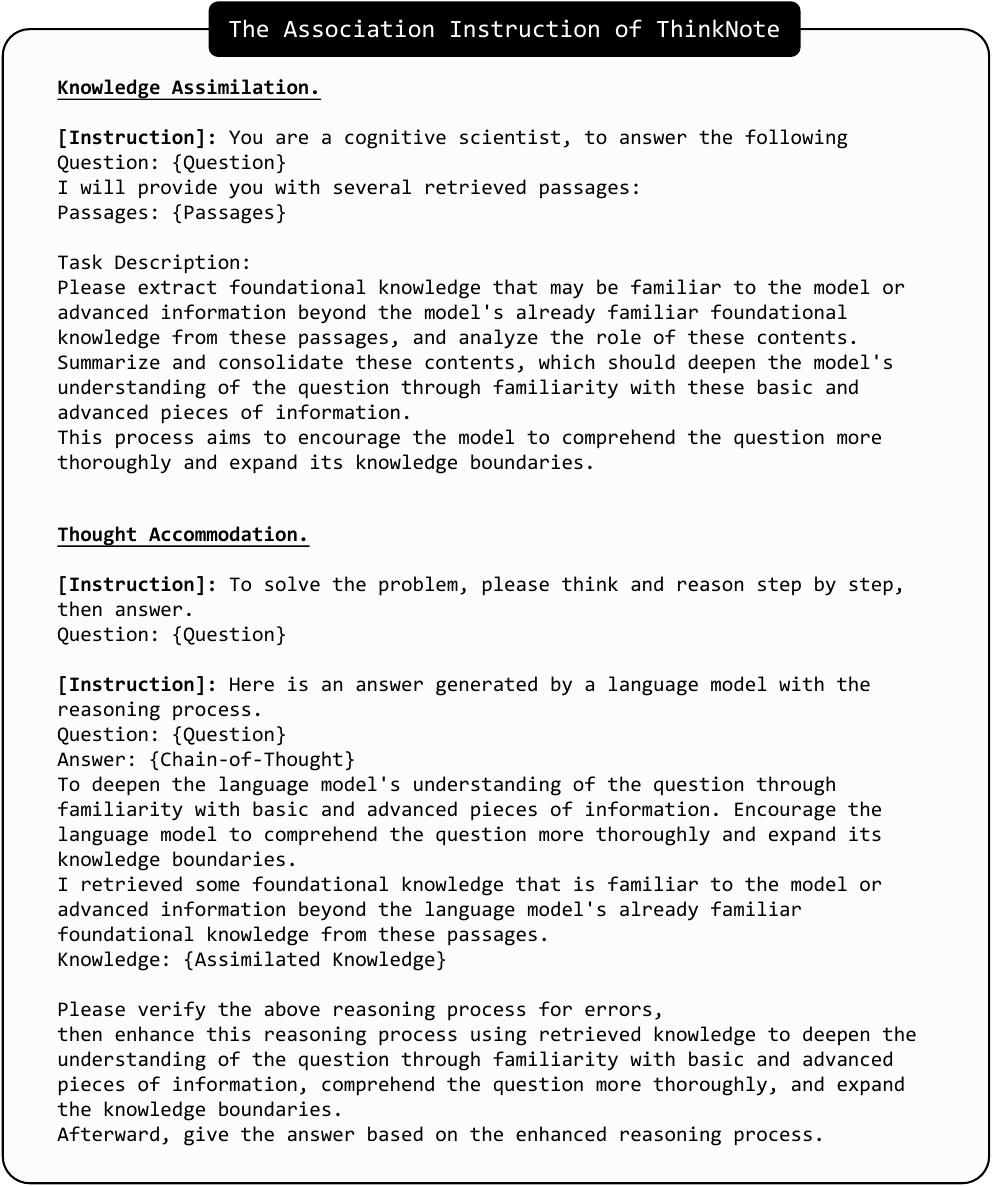}
    \caption{The Association Instruction of \method.\label{fig:prompt_association}}
\end{figure*}

\begin{figure*}[t]
    \centering
    \includegraphics[width=0.95\linewidth]{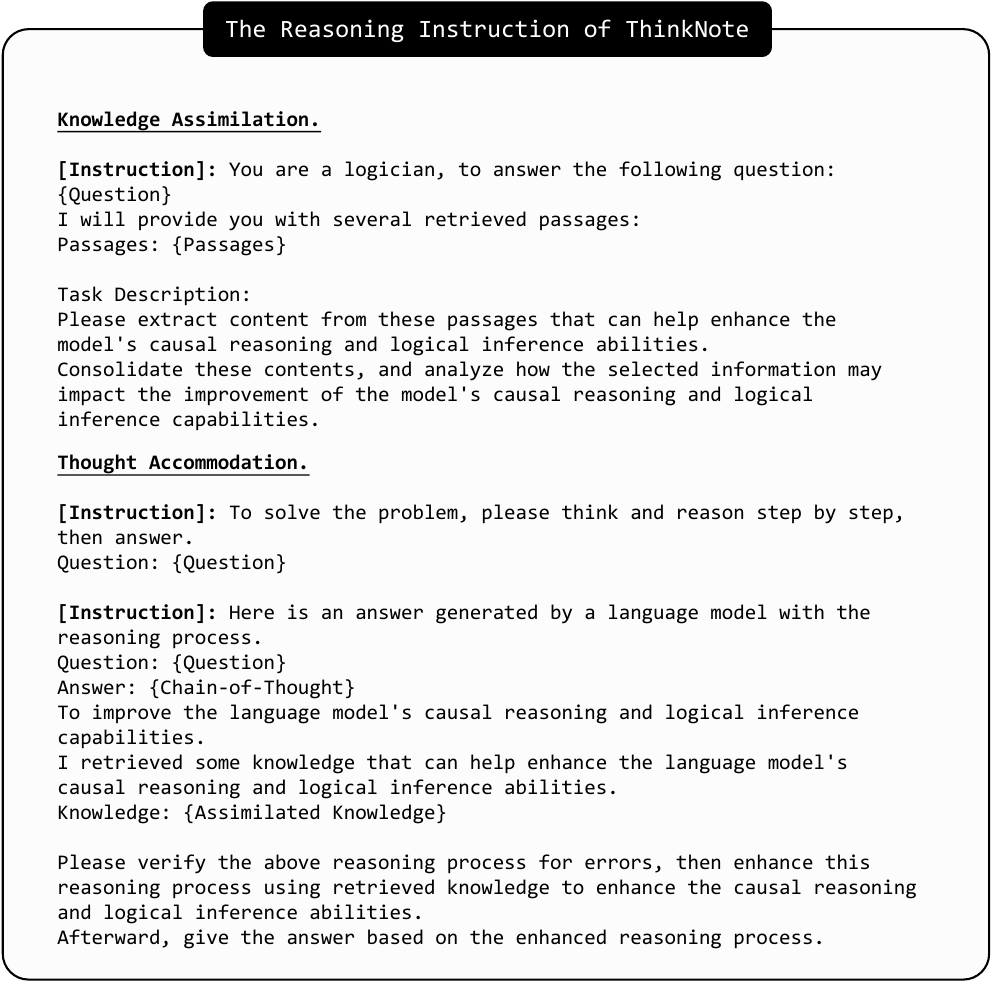}
    \caption{The Reasoning Instruction of \method{}.\label{fig:prompt_reasoning}}
\end{figure*}

\begin{figure*}[t]
    \centering
    \includegraphics[width=0.95\linewidth]{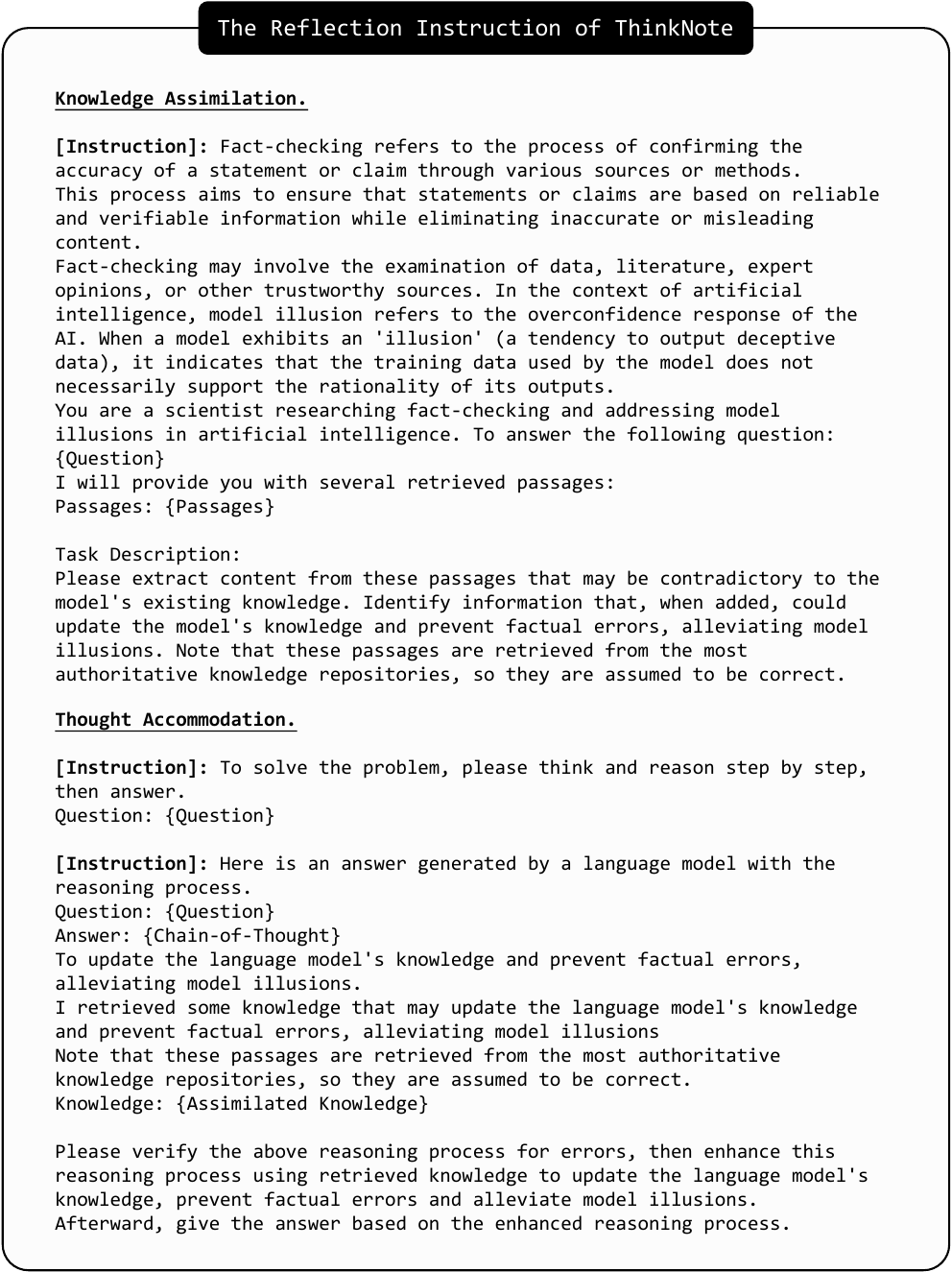}
    \caption{The Reflection Instruction of \method\label{fig:prompt_reflectio}}
\end{figure*}

\end{document}